\documentclass[12pt,journal,draftcls,letterpaper,onecolumn]{IEEEtran}
\usepackage{amssymb}
\usepackage{graphicx,subfigure}
\usepackage{subfigure}
\usepackage{algorithmic}
\usepackage{algorithm}
\usepackage{setspace}

\newtheorem{mythm}{Theorem}
\newtheorem{mycoly}{Corollary}
\newtheorem{mylm}{Lemma}
\newtheorem{mypropos}{Proposition}

\usepackage{amsmath}
\usepackage{fmtcount}
\usepackage{color}
\begin{document}
%
\title{Asymptotic Generalization Bound of Fisher's Linear Discriminant Analysis}
%
%
%

\author{Wei~Bian 
        and~Dacheng~Tao,~\IEEEmembership{Senior Member,~IEEE}
\thanks{}
\thanks{The authors are with the Centre for Quantum Computation and Intelligent Systems, Faculty of Engineering and Information Technology, University of Technology, Sydney, Australia.
E-mail: wei.bian@student.uts.edu.au
 and dacheng.tao@uts.edu.au}}

\maketitle

\begin{abstract}
Fisher's linear discriminant analysis (FLDA) is an important dimension reduction method in statistical pattern recognition. It has been shown that FLDA is asymptotically Bayes optimal under the homoscedastic Gaussian assumption. However, this classical result has the following two major limitations: 1) it holds only for a fixed dimensionality $D$, and thus does not apply when $D$ and the training sample size $N$ are proportionally large; 2) it does not provide a quantitative description on how the generalization ability of FLDA is affected by $D$ and $N$. In this paper, we present an asymptotic generalization analysis of FLDA based on random matrix theory, in a setting where both $D$ and $N$ increase and $D/N\longrightarrow\gamma\in[0,1)$. The obtained lower bound of the generalization discrimination power overcomes both limitations of the classical result, i.e., it is applicable when $D$ and $N$ are proportionally large and provides a quantitative description of the generalization ability of FLDA in terms of the ratio $\gamma=D/N$ and the population discrimination power. Besides, the discrimination power bound also leads to an upper bound on the generalization error of binary-classification with FLDA.
\end{abstract}


%

\section{Introduction}
%
%
%
%
Fisher's linear discriminant analysis (FLDA) \cite{Fisher36} \cite{Rao48} is one of the most representative dimension reduction techniques in statistical pattern recognition . By projecting examples into a low dimensional subspace with maximum discrimination power, FLDA helps improve the accuracy and the robustness of a decision system \cite{Loog01TPAMI} \cite{TaoGeometricmean} \cite{MaxMinDA} \cite{ODA}. During the past decades, FLDA has been applied to a wide range of areas, from speech/music classification \cite{speech_lda} \cite{alexandre05}, face recognition \cite{Belhumeur} \cite{KimK05} to financial data analysis \cite{Altman68} \cite{Kumar06}.

An important property of FLDA is its asymptotic Bayes optimality under the homoscedastic Gaussian assumption \cite{BLDA_reference} \cite{Bickel_Levina04} \cite{Fan_Fan_Wu11} , which is a corollary of classical results from multivariate statistics \cite{Anderson84}. Actually, as training sample size $N$ goes to infinity, both the within-class scatter matrix $\widehat{\boldsymbol\Sigma}$ (sample covariance) and the between-class scatter matrix $\widehat{\mathbf S}$ converge to their population counterparts $\boldsymbol\Sigma$ and $\mathbf S$. Therefore, the empirically optimal projection matrix $\widehat{\mathbf W}$ of FLDA, obtained by generalized eigendecomposition over $\widehat{\boldsymbol\Sigma}$ and $\widehat{\mathbf S}$, also converges to its population counterpart $\mathbf W$. Thanks to the asymptotic Bayes optimality, we can expect an acceptable performance of FLDA as long as $N$ is sufficiently large. However, this classical result, i.e., the asymptotic Bayes optimality, suffers from two major limitations:
\begin{enumerate}
  \item It is obtained by fixing the dimensionality $D$ and letting only $N$ increase to infinity. But in practice, $D$ and $N$ can be proportionally large, which makes the classical result inapplicable.
  \item It does not provide quantitative description on the performance of FLDA, especially, how the generalization ability of FLDA is affected by $D$ and $N$.
\end{enumerate}

\subsection{The Contribution of this Paper}
To address aforementioned limitations of the classical result, in this paper, we present an asymptotic generalization analysis of FLDA. Our analysis is superior from two aspects. First, we modify the setting of analysis by allowing both $D$ and $N$ to increase and assuming the dimensionality to training sample size ratio $\gamma=D/N$ has a limit in $[0,1)$. This makes our result applicable in the case where $D$ and $N$ are proportionally large. Second, we quantitatively examine the generalization ability of FLDA. Denoting by $\Delta(\boldsymbol\Sigma,\mathbf S|\widehat{\mathbf W})$ the generalization discrimination power of FLDA, we intend to bound it from the lower side in terms of $D$ and $N$, with respect to the population discrimination power  $\Delta(\boldsymbol\Sigma,\mathbf S|\mathbf W)$. Taking a binary-class problem, for example: suppose $\Delta(\boldsymbol\Sigma,\mathbf S|\mathbf W)=\boldsymbol\lambda$ and $\gamma=D/N$, then our asymptotic generalization bound shows that $\Delta(\boldsymbol\Sigma,\mathbf S|\widehat{\mathbf W})$ is almost surely larger than
\begin{equation}\notag
    \cos^2(\arccos(\sqrt{{\boldsymbol\lambda}/(\boldsymbol\lambda + \gamma)}) + \arccos(\sqrt{1-\gamma}))\boldsymbol\lambda,
\end{equation}
under mild conditions. Further, as a corollary of the discrimination power bound, we also obtain an asymptotic generalization error bound for binary classification with FLDA.

Based on the obtained asymptotic generalization bound, we can get better insight of FLDA. It is commonly known that the performance of covariance estimation has a severe influence to the generalization ability of FLDA. By assuming a sufficient population discrimination power so as to eliminate the influence from between-class matrix estimation, we show that the mere influence from covariance estimation is proportional to the ratio $\gamma=D/N<1$, i.e., due to the imperfection of covariance estimation, $\Delta(\boldsymbol\Sigma,\mathbf S|\widehat{\mathbf W})$ is about $1-\gamma$ times of $\Delta(\boldsymbol\Sigma,\mathbf S|\mathbf W)$. It is worth noticing that such result holds independent of the covariance $\boldsymbol\Sigma$. Besides, the bound shows that the performance of FLDA is substantially determined by the ratio $\gamma=D/N$, given a fixed population discrimination power $\Delta(\boldsymbol\Sigma,\mathbf S|{\mathbf W})$. Therefore, $N$ only needs to scale linearly with respect to $D$ for an acceptable generalization ability of FLDA, although a quadratic number of parameters are to be estimated in the sample covariance.

\subsection{Tools}
The technical tools used in our asymptotic generalization analysis are from random matrix theory (RMT) \cite{wigner1955characteristic} \cite{wigner1958distribution} \cite{marcenko1967distribution} \cite{edelman2005random}, the main goal of which is to provide understanding of the statistics of eigenvalues of matrices with entries drawn randomly from various probability distributions. RMT was originally motivated by applications in nuclear physics in 1950's, and then it was intensively studied in mathematics and statistics. It also found successful applications in engineering fields, e.g., wireless communications \cite{tulino2004random}, recently. In this paper, we make use of two important results from RMT. The first one is the Mar{\v{c}}enko-Pastur Law \cite{marcenko1967distribution}, which states that the empirical spectral distribution of a Wishart random matrix converges almost surely to a deterministic distribution $F_{\gamma}(\lambda)$ as $\lim\gamma=D/N\in[0,1)$. The second one is the almost sure convergence of the extreme singular values of a large Gaussian random matrix \cite{edelman2005random}. We formulate these two results in following propositions.

\begin{mypropos}\label{propos:MP_Law1}
Given $\mathbf G\in\mathbb R^{D\times N}$, whose entries are independently sampled from standard Gaussian distribution $\mathcal N(0,1)$, then as both $D$ and $N\longrightarrow\infty$ and $D/N\longrightarrow\gamma\in[0,1)$, the empirical distribution of the eigenvalues of $\frac1N\mathbf G\mathbf G^T$, i.e.,
\begin{equation}
  F_N(\lambda) = \frac1D\sum_{i=1}^D1\big\{\lambda_i\big(\frac1N\mathbf G\mathbf G^T\big)\le\lambda\big\},~\lambda\ge0,
\end{equation}
converges almost surely to a deterministic limit distribution $F_{\gamma}(\lambda)$ with density
\begin{equation}
    dF_{\gamma}(\lambda)= \frac{\sqrt{(\lambda_+ - \lambda)(\lambda-\lambda_-)}}{2\pi\gamma\lambda}d\lambda,
\end{equation}
where
\begin{equation}
    \lambda_+ = (1+\sqrt{\gamma})^2 \mbox{~and~} \lambda_- = (1-\sqrt{\gamma})^2.
\end{equation}
\end{mypropos}

\begin{mypropos}\label{lm:largest_singularvalue_GRM}
Letting $\mathbf G\in\mathbb R^{D\times m}$ with i.i.d. entries sampled from $\mathcal N(0,1)$, then as $m/D\longrightarrow\gamma\in[0,1)$,
\begin{equation}
    \frac{1}{\sqrt{D}}\sigma_{max}(\mathbf G)\overset{a.s.}\longrightarrow1+\sqrt{\gamma},
\end{equation}
and
\begin{equation}
    \frac{1}{\sqrt{D}}\sigma_{min}(\mathbf G)\overset{a.s.}\longrightarrow1-\sqrt{\gamma}.
\end{equation}
\end{mypropos}

\subsection{Notations}
Throughout this paper, we will use the following notations. Bold lower case letter $\mathbf a$ denotes a vector. Bold upper case letter $\mathbf A$ denotes a matrix. $\mathbb R^D$ denotes a $D$-dimensional vector space. $\mathbb R^{D_1\times D_2}$ denotes the set of all $D_1$ by $D_2$ matrices. $\mathbf A_{ii}$ or $\{\mathbf A\}_{ii}$ denotes the $i$-th diagonal entry of a symmetric matrix $\mathbf A$. $\mathbf A_i$ denotes the $i$-th column of $\mathbf A$. $\mathbf A_{1:c}$ denotes the matrix composed by the first $c$ columns of $\mathbf A$. $\mathbb S^{D-1}$ denotes the $D$-dimensional unit sphere located on the original point. $\mathbb S_{++}^{D\times D}$ denotes the set of all $D$ by $D$ positive definite matrices. $\|\mathbf a\|$ denotes the $\ell_2$ norm of $\mathbf a$. $\sigma_{max}(\mathbf A)$ and $\sigma_{min}(\mathbf A)$ are the extreme singular values of $\mathbf A$. $\|\mathbf A\|=\sigma_{max}(\mathbf A)$ denotes the operator norm of $\mathbf A$. $\lambda_i(\mathbf A)$ denotes the $i$-th eigenvalue of $\mathbf A$, sorted in a descent order. $\Lambda(\mathbf A)$ denotes the diagonal matrix composed of the eigenvalues of $\mathbf A$, with the eigenvalues sorted in a descent order. $\mathcal R(\mathbf A)$ denotes an orthogonal basis of the range or the column space of $\mathbf A$. $[\mathbf e_1,...,\mathbf e_D]$ is the canonical basis of $\mathbf R^D$.

\section{Main Result}\label{sec:preliminary}

\subsection{Bounding Generalization Discrimination Power}
Suppose we have $c+1$ classes, represented by homoscedastic Gaussian distributions in a high-dimensional space $\mathbb R^D$, $\mathcal N_i(\boldsymbol\mu_i,\boldsymbol\Sigma)$, $i=1,2,...,c+1$, with class means $\boldsymbol\mu_i\in\mathbb R^D$ and the common covariance matrix $\boldsymbol\Sigma\in\mathbb S^{D\times D}_{++}$. Assuming the classes have equal prior probability $\frac1{c+1}$\footnote{For the convenience of expression, we assume an equal prior probability. This does not substantially change the analysis throughout this paper.}, the following matrix $\mathbf S$, which is referred to as the between-class scatter matrix, gives a measure of class separation,
\begin{equation}\label{eq:betweenScatter_S}
    \mathbf S = \frac1{c+1}\sum_{i=1}^{c+1}(\boldsymbol\mu_i-\boldsymbol\mu)(\boldsymbol\mu_i-\boldsymbol\mu)^T, \mbox{~with~} \boldsymbol\mu=\frac1{c+1}\sum_{i=1}^{c+1}\boldsymbol\mu_i.
\end{equation}
Suppose the eigendecomposition of $\mathbf \Sigma^{-1}\mathbf S$ has (at most) $c$ nonzero eigenvalues $\boldsymbol\lambda_i$, $i=1,2,...,c$, and associated eigenvectors $\mathbf W=[\mathbf w_1,..,\mathbf w_c]$. FLDA uses $\mathbf W$ as a projection matrix to obtain a low-dimensional data representation, and according to Fisher's criterion, the discrimination power in the dimension reduced space is given by \cite{FukunagaBook}
\begin{equation}\label{eq:disc_power}
    \Delta(\boldsymbol\Sigma,\mathbf S|\mathbf W) = \mathrm{Tr}\left((\mathbf W^T\boldsymbol\Sigma\mathbf W)^{-1}\mathbf W^T\mathbf S\mathbf W\right)=\sum_{i=1}^c\boldsymbol\lambda_i.
\end{equation}

In practice, we do not have access to population parameters ${\boldsymbol\Sigma}$ and ${\mathbf S}$, but their estimates, i.e., the sample covariance $\widehat{\boldsymbol\Sigma}$ and the sample between-class scatter matrix $\widehat{\mathbf S}$ via sample class means $\widehat{\boldsymbol\mu}_i$. Denoting by $\widehat{\mathbf W}$ the empirical projection matrix obtained from generalized eigendecomposition of $\widehat{\boldsymbol\Sigma}$ and $\widehat{\mathbf S}$, the generalization discrimination power of FLDA is given by
\begin{equation}\label{eq:generalization_disc_power}
    \Delta(\boldsymbol\Sigma,\mathbf S|\widehat{\mathbf W}) = \mathrm{Tr}\left((\widehat{\mathbf W}^T\boldsymbol\Sigma\widehat{\mathbf W})^{-1}\widehat{\mathbf W}^T\mathbf S\widehat{\mathbf W}\right),
\end{equation}
which measures how the classes are separated in the dimension reduced space. When data dimensionality $D$ is fixed and training sample size $N$ goes to infinity, the generalization discrimination power (\ref{eq:generalization_disc_power}) will converge to its population counterpart (\ref{eq:disc_power}), since $\widehat{\mathbf W}$ converges to $\mathbf W$. However, such classical result is invalid when $D$ increases proportionally with $N$. Regarding this, the following theorem gives a new asymptotic result on FLDA's generalization ability, in a setting where $D$ and $N$ increase to infinity proportionally.

\begin{mythm}\label{thm:generalization_bound}
Suppose the population discrimination power $\Delta(\boldsymbol\Sigma,\mathbf S|\mathbf W)=\sum_{i=1}^c\boldsymbol\lambda_i$. The generalization discrimination power $\Delta(\boldsymbol\Sigma,\mathbf S|\widehat{\mathbf W})$ can be factorized as
\begin{align}
 \Delta(\boldsymbol\Sigma,\mathbf S|\widehat{\mathbf W}) =\sum_{i=1}^c\boldsymbol\delta_i\boldsymbol\lambda_i
\end{align}
where $0\le \boldsymbol\delta_i\le1$. Further, as both the dimensionality $D$ and the training sample size $N$ increase ($N>D$) and $D/N\longrightarrow\gamma\in[0,1)$, it holds asymptotically
\begin{equation}\label{eq:weight_factor}
    \boldsymbol\delta_i\boldsymbol\lambda_i\ge{\max}^2\big\{\cos(\arccos(\sqrt{{\boldsymbol\lambda_i}/(\boldsymbol\lambda_i + \gamma)}) + \arccos(\sqrt{1-\gamma})),0\big\}\boldsymbol\lambda_i, ~a.s.
\end{equation}
\end{mythm}

\begin{figure}
\centering
\subfigure[Lower Bound of Discrimination Power]{\includegraphics[width=0.495\columnwidth]{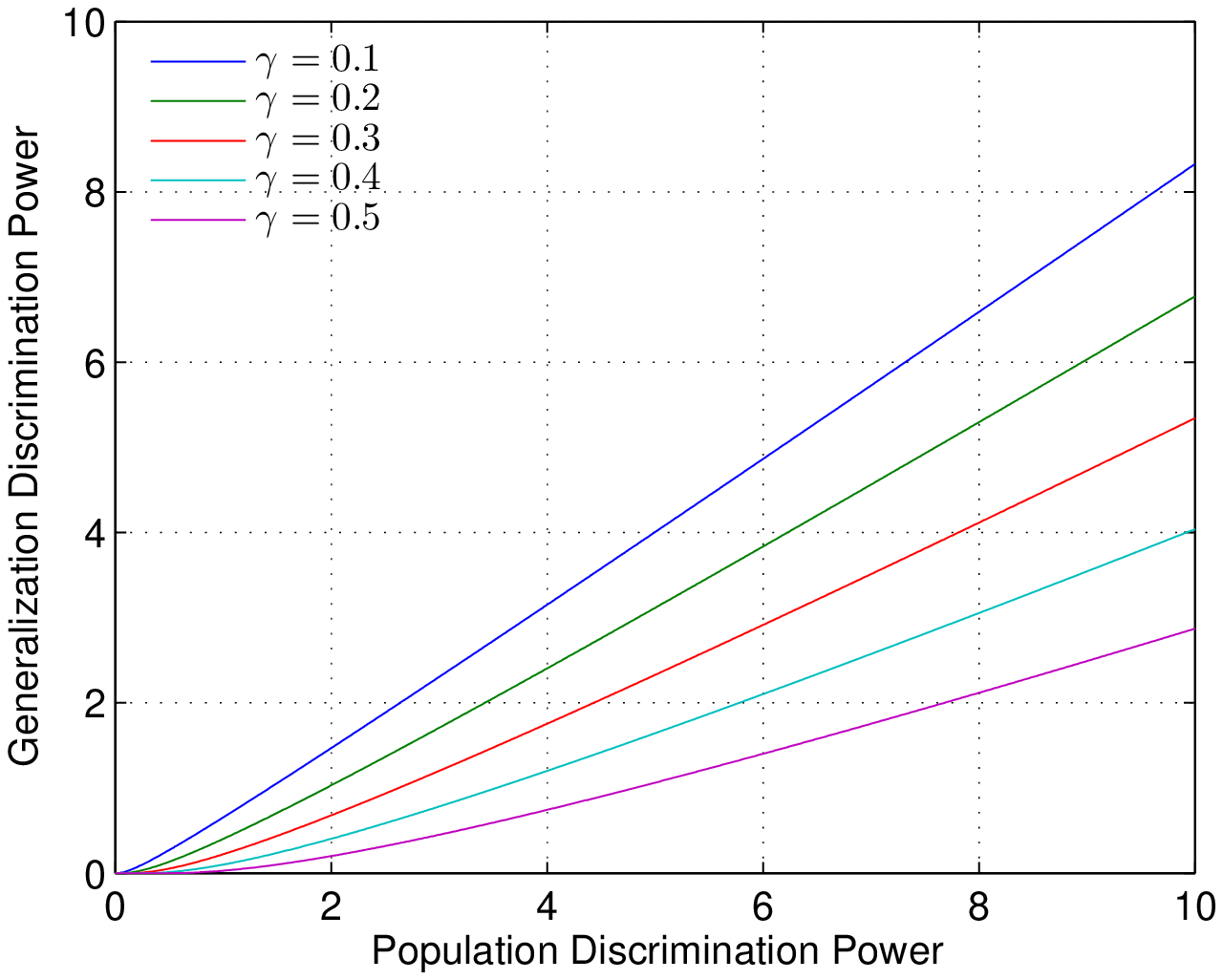}}
\subfigure[Upper Bound of Binary Classification Error]{\includegraphics[width=0.495\columnwidth]{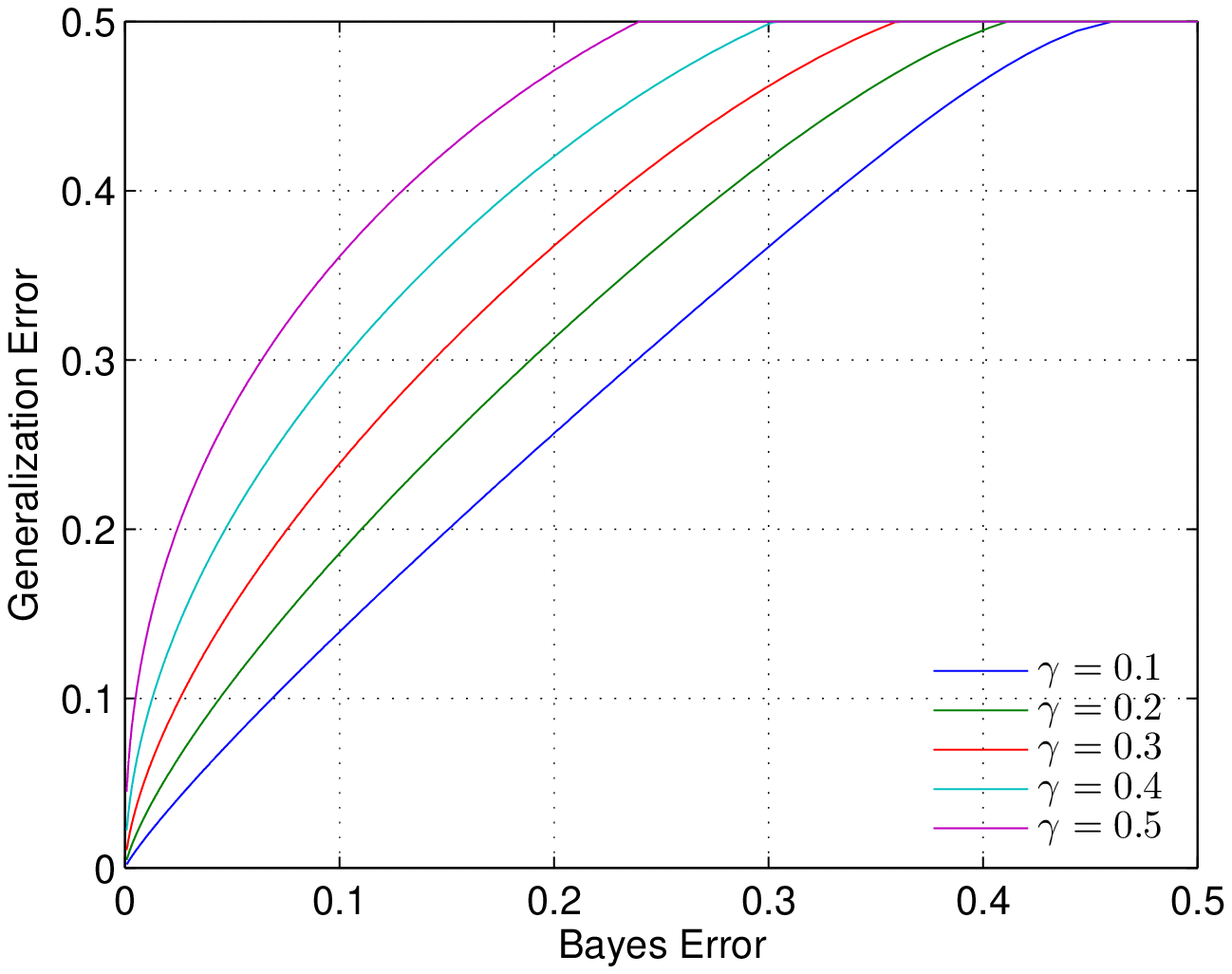}}
\caption{Asymptotic Generalization Bound of Fisher's Linear Discriminant Analysis.}\label{fig:generalization_bound_power_error}
\end{figure}

Theorem \ref{thm:generalization_bound} gives an asymptotically lower bound on the generalization ability of FLDA, in terms of the population discrimination power $\boldsymbol\lambda_i$ and the dimensionality to training sample size ratio $\gamma=D/N$.  An important feature of the bound is that it is determined by the ratio $\gamma=D/N$ rather than the dimensionality D. In other words, a good generalization performance of FLDA only requires a training sample size that scales linearly with respect the dimensionality, although there are a quadratic number of parameters to be estimated in the sample covariance. Figure \ref{fig:generalization_bound_power_error} (a) gives an illustration of the bound under different values of the ratio $\gamma=D/N$.

Besides, according to (\ref{eq:weight_factor}), the influence of the ratio $\gamma=D/N$ to the lower bound comes from two aspects, each through the term $\sqrt{\boldsymbol\lambda_i/(\boldsymbol\lambda_i + \gamma)}$ and the term $\sqrt{1-\gamma}$. Note that  $\sqrt{\boldsymbol\lambda_i/(\boldsymbol\lambda_i + \gamma)}$ allows a tradeoff between $\boldsymbol\lambda_i$ and $\gamma$, i.e., when $\boldsymbol\lambda_i$ is sufficiently large, $\arccos(\sqrt{{\boldsymbol\lambda_i}/(\boldsymbol\lambda_i + \gamma)})$ approaches $0$ and thus vanishes from the lower bound (\ref{eq:weight_factor}). The second term $\sqrt{1-\gamma}$ only depends on $\gamma$, and later proofs reveal that it measures how covariance estimation influences the generalization of FLDA. Assuming a sufficient large $\boldsymbol\lambda_i$ such that $\sqrt{\boldsymbol\lambda_i/(\boldsymbol\lambda_i + \gamma)}\approx1$, we have
\begin{equation}\label{eq:cov_lda}
       \boldsymbol\delta_i\boldsymbol\lambda_i \approx (1-\gamma)\boldsymbol\lambda_i,
\end{equation}
which shows that the loss of discrimination power due to the imperfection of covariance estimation is approximately proportion to $\gamma$. To the best of our knowledge, this is the simplest quantitative result on the influence of covariance estimation to FLDA, compared with related studies in the literature \cite{Bickel_Levina04} \cite{Durrant10_ICPR_FLDA} \cite{Hoyle_2011_pami}. It is worth noticing that, as long as $\boldsymbol\Sigma\in\mathbb S_{++}^{D\times D}$, the result is independent of the spectrum of the population covariance $\boldsymbol\Sigma$, e.g., the extreme eigenvalues $\lambda_{min}(\boldsymbol\Sigma)$ and $\lambda_{max}(\boldsymbol\Sigma)$,  or the conditional number $\lambda_{max}(\boldsymbol\Sigma)/\lambda_{min}(\boldsymbol\Sigma)$.

\subsection{Bounding Generalization Error of Binary Classification}
In binary-class case, FLDA can also be regarded as a linear classifier, where the hyperplane of the linear classifier is perpendicular to the one-dimensional projection vector $\widehat{\mathbf w}_1$ of dimension reduction. Without loss of generality, suppose $\widehat{\mathbf w}_1^T(\boldsymbol\mu_1-\boldsymbol\mu_2)\ge0$, the generalization error $P$ of binary classification with FLDA can be calculated analytically by \cite{Mclachlan_PR}
\begin{align}
  P =  0.5\Phi \left\{-\frac{\widehat{\mathbf w}_1^T\boldsymbol\mu_1-0.5\widehat{\mathbf w}_1^T(\widehat{\boldsymbol\mu}_1+\widehat{\boldsymbol\mu}_2)}{\sqrt{\widehat{\mathbf w}_1^T\boldsymbol\Sigma\widehat{\mathbf w}_1}}\right\}+0.5\Phi \left\{-\frac{0.5\widehat{\mathbf w}_1^T(\widehat{\boldsymbol\mu}_1+\widehat{\boldsymbol\mu}_2)-\widehat{\mathbf w}_1^T\boldsymbol\mu_2}{\sqrt{\widehat{\mathbf w}_1^T\boldsymbol\Sigma\widehat{\mathbf w}_1}}\right\},\label{eq:flda_gener_error1}
\end{align}where $\Phi(\cdot)$ is the cumulative distribution function (CDF) of the standard Gaussian. If we replace $\widehat{\mathbf w}_1$ and $\widehat{\boldsymbol\mu_i}$ by its population counterpart $\mathbf w_1$ and $\boldsymbol\mu_i$, then (\ref{eq:flda_gener_error1}) gives the Bayes error $P_{Bayes}$, i.e.,
\begin{align}
  P_{Bayes} = \Phi\left\{-\frac{0.5\mathbf w_1^T(\boldsymbol\mu_1-\boldsymbol\mu_2)}{\sqrt{\mathbf w_1^T\boldsymbol\Sigma\mathbf w_1}}\right\}=\Phi\left\{-\sqrt{\frac{\mathbf w_1^T\mathbf S\mathbf w_1}{\mathbf w_1^T\boldsymbol\Sigma\mathbf w_1}}\right\}=\Phi\left(-\sqrt{\boldsymbol\lambda_1}\right).
\end{align}
Below, we present a corollary of Theorem \ref{thm:generalization_bound}, which gives an asymptotic upper bound of $P$ in terms of $P_{Bayes}$ and $\gamma=D/N$.

\begin{mycoly}\label{coly:generalization_bound}
For binary classification with equal prior probabilities, suppose the population discrimination power $\Delta(\boldsymbol\Sigma,\mathbf S|\mathbf w_1)=\boldsymbol\lambda_1$, then if both dimensionality $D$ and training sample size $N$ increase ($N>D$) and $D/N\longrightarrow\gamma\in[0,1)$, the generalization error $P$ of FLDA can be upper bounded asymptotically by
\begin{align}\label{eq:generalization_error}
 P\le\Phi\left(-\varrho\sqrt{\boldsymbol\lambda_1}\right), ~a.s.
\end{align}
where
\begin{equation}\label{eq:varrho}
    \varrho={\max}\big\{\cos(\arccos(\sqrt{{\boldsymbol\lambda_1}/(\boldsymbol\lambda_1 + \gamma)}) + \arccos(\sqrt{1-\gamma})),0\big\}.
\end{equation}
Further since the Bayes error $P_{Bayes}=\Phi\left(-\sqrt{\boldsymbol\lambda_1}\right)$, it holds asymptotically
\begin{align}
  P\le\Phi\big(\varrho\Phi^{-1}\left(P_{Bayes}\right)\big),~a.s.
\end{align}
with
\begin{equation}\label{eq:varrho}
    \varrho={\max}\left\{\cos\left(\arccos\left(\sqrt{\frac{(\Phi^{-1}(P_{Bayes}))^2}{((\Phi^{-1}(P_{Bayes}))^2 + \gamma}}\right) + \arccos(\sqrt{1-\gamma})\right),0\right\}.
\end{equation}
\end{mycoly}

Similar to the discrimination power bound, Corollary \ref{coly:generalization_bound} shows that, given a binary classification problem with Bayes error $P_{Bayes}$, the generalization error of FLDA is also determined by the dimensionality to training sample size ratio $\gamma=D/N$. Figure \ref{fig:generalization_bound_power_error} (b) gives an illustration of the generalization error bound under different values of $\gamma$.

\subsection{Related Work}

In recent years, asymptotic analysis on FLDA have also been performed in the case where $D>N$. For example, \cite{Bickel_Levina04} found that when $D$ increases faster than $N$ the the pseudo-inverse based FLDA approaches to a random guess and therefore suggested a ``naive Bayes'' approach in this situation. A more detailed analysis on pseudo-inverse FLDA was given in \cite{Hoyle_2011_pami} by investigating the estimation error of pseudo-inverse of the sample covariance. Random matrix theory, e.g., Mar\v cenko-Pastur Law, was also utilized in \cite{Hoyle_2011_pami}, so as to bound the expected estimation error in the asymptotic case. The result in this paper provides a complementary theory of FLDA in the setting of $D<N$, which shows that the generalization ability of FLDA in such situation is mainly determined by the ratio $\gamma=D/N$.

In contrast to asymptotic analysis, generalization bounds in finite sample case were derived most recently in both linear and kernel spaces, and by using random projection as regularization if $D>N$  \cite{Durrant10_ICPR_FLDA} \cite{Durrant10_KDD_FLDA} \cite{Durrant_KFLDA}. The advantage of these results is they provide explicit probability bounds for finite $N$ and $D$, while asymptotic results inherently require sufficient large $N$ and $D$. However, we would like to emphasize that the bounds obtained in this paper have their own merit, by linking the generalization discrimination power (or generalization error) to the population discrimination power (or Bayes error) directly in terms of the ratio $\gamma=D/N$. Besides, as shown by empirical evaluation in later section IV, the bounds hold with high probability (in the empirical sense) for moderate $D$ and $N$, though they are obtained asymptotically.

\section{Proof of Main Result}
In this section, we present the proof of Theorem \ref{thm:generalization_bound}, which are mainly based upon the asymptotic results on eigensystems of the sample covariance and the sample between-class scatter matrix.
\subsection{On $\Delta(\boldsymbol\Sigma,\mathbf S|\widehat{\mathbf W})$}
We begin the proof by bounding the generalization discrimination power $\Delta(\boldsymbol\Sigma,\mathbf S|\widehat{\mathbf W})$ in terms of eigenvalues and/or eigenvectors of a normalized version of the sample covariance and sample between-class scatter matrix.

\begin{mylm}\label{lm:double_diagonal}
Given a problem with population discrimination power $\Delta(\boldsymbol\Sigma,\mathbf S|\mathbf W)=\sum_{i=1}^c\boldsymbol\lambda_i$, there is a nonsingular matrix $\mathbf X$ that simultaneously diagonalizes $\boldsymbol\Sigma$ and $\mathbf S$, i.e.,
\begin{equation}\label{eq:doule_diagonal}
\mathbf X^{T}{\boldsymbol \Sigma}\mathbf X =\mathbf I\mbox{~and~} \mathbf X^{T}{\mathbf S}\mathbf X = \boldsymbol\Lambda_0,
\end{equation}
where $\boldsymbol\Lambda_0=\mbox{diag}(\boldsymbol\lambda_1,...,\boldsymbol\lambda_c,0,...,0)$.
\end{mylm}

\begin{mylm}\label{lm:gen_dis_pow} Given the normalized estimates $\widehat{\boldsymbol \Sigma}_0 = \mathbf X^{T}\widehat{\boldsymbol \Sigma}\mathbf X$ and $\widehat{\mathbf S}_0 =  \mathbf X^{T}\widehat{\mathbf S}\mathbf X$, and their eigendecompositions $\widehat{\boldsymbol\Sigma}_0=\mathbf U\Lambda(\widehat{\boldsymbol\Sigma}_0)\mathbf U^T$ and $\widehat{\mathbf S}_0=\mathbf V\Lambda(\widehat{\mathbf S}_0)\mathbf V^T$, the generalization discrimination power $\Delta(\boldsymbol\Sigma,\mathbf S|\widehat{\mathbf W})$ can be expressed as
\begin{equation}\label{eq:general_dis_power_delta}
    \Delta(\boldsymbol\Sigma,\mathbf S|\widehat{\mathbf W})=\sum_{i=1}^c{\boldsymbol\delta_i\boldsymbol\lambda_i},
\end{equation}
where
\begin{equation}\label{eq:delta}
    \boldsymbol\delta_i=\big\|\mathcal R^T\Big(\Lambda^{-1}(\widehat{\boldsymbol\Sigma}_0)\mathbf U^T\mathbf V_{1:c}\Big)\mathbf U^T\mathbf e_i\big\|^2.
\end{equation}
\end{mylm}

\begin{mylm}\label{lm:lowerbound_delta}
Given $\Lambda(\widehat{\boldsymbol\Sigma}_0)$ and $\mathbf V_{1:c}$ from Lemma \ref{lm:gen_dis_pow}, it holds
\begin{align}\label{eq:lower_bound_delta}
    \boldsymbol\delta_i\ge{\max}^2\Big\{\cos\left(\arccos(\|\mathbf V_{1:c}^T\mathbf e_i\|) + \arccos\left(\xi^T\Lambda^{-1}(\widehat{\boldsymbol\Sigma}_0)\xi\Big/
    \sqrt{\xi^T\Lambda^{-2}(\widehat{\boldsymbol\Sigma}_0)\xi}\right)\right),0\Big\}.
\end{align}
where $\xi$ is a unit-length random vector uniformly distributed on the unit sphere $\mathbb S^{D-1}$.
\end{mylm}

\quad\\
\indent Lemma \ref{lm:gen_dis_pow} and Lemma \ref{lm:lowerbound_delta} show that the generalization discrimination power of FLDA are determined by the eigensystems of the normalized estimates $\widehat{\boldsymbol\Sigma}_0$ and $\widehat{\mathbf S}_0$. Since $\widehat{\boldsymbol\Sigma}_0$ is actually an estimate of the identity covariance matrix $\mathbf I$, we have that given the population discrimination power $\Delta(\boldsymbol\Sigma,\mathbf S|\mathbf W)=\sum_{i=1}^c\boldsymbol\lambda_i$, the generalization ability of FLDA, i.e., $\Delta(\boldsymbol\Sigma,\mathbf S|\widehat{\mathbf W})=\sum_{i=1}^c\boldsymbol\delta_i\boldsymbol\lambda_i$, is independent of the population covariance $\boldsymbol\Sigma$. Next, we present properties on the eigensymstems of $\widehat{\boldsymbol\Sigma}_0$ and $\widehat{\mathbf S}_0$, which are necessary for evaluating the lower bound of $\boldsymbol\delta_i$ in (\ref{eq:lower_bound_delta}).

\subsection{Properties of $\widehat{\boldsymbol\Sigma}_0$}
We have the following lemma on the eigensystem of the normalized sample covariance $\widehat{\boldsymbol\Sigma}_0$.

\begin{mylm}\label{lm:property_Sigma0}
Given the eigendecomposition $\widehat{\boldsymbol\Sigma}_0=\mathbf U\Lambda(\widehat{\boldsymbol\Sigma}_0)\mathbf U^T$, it holds
\begin{enumerate}
  \item $\mathbf U$ and $\Lambda(\widehat{\boldsymbol\Sigma}_0)$ are independent random variables;
  \item $\mathbf U$ follows the Haar distribution, i.e., it is uniformly distributed on the set of all orthonormal matrices in $\mathbb R^{D\times D}$;
  \item denoting by $F_N(\lambda)$ the empirical spectral distribution of the eigenvalues of $\widehat{\boldsymbol\Sigma}_0$, i.e.,
  \begin{equation}\label{eq:EMF}
    F_N(\lambda) = \frac1D\sum_{i=1}^D1\{\lambda_i(\widehat{\boldsymbol\Sigma}_0)\le\lambda\},~\lambda\ge0,
  \end{equation}
  then, as $D/N\longrightarrow\gamma\in[0,1)$,
  \begin{equation}\label{eq:MPlaw}
    F_N(\lambda)\overset{a.s.}\longrightarrow F_{\gamma}(\lambda),
  \end{equation}
  where the limit distribution $F_{\gamma}(\lambda)$ has the density
  \begin{equation}\label{eq:density}
    dF_{\gamma}(\lambda) = \frac1{2\pi\gamma}\frac{\sqrt{(\lambda_+-\lambda)(\lambda-\lambda_-)}}{\lambda}d\lambda,
  \end{equation}
  with
  \begin{equation}\label{eq:boundary}
    \lambda_+ = (1+\sqrt{\gamma})^2\mbox{~and~}\lambda_-= (1-\sqrt{\gamma})^2.
  \end{equation}
\end{enumerate}
\end{mylm}

The first and the second statements in Lemma \ref{lm:property_Sigma0} can be understood by the fact that $\widehat{\boldsymbol\Sigma}_0$ is an empirical estimate of $\mathbf I$, whose probability density is invariant to any orthogonal transformation. The last statement is a corollary of the Mar{\v{c}}enko-Pastur law, i.e., Proposition \ref{propos:MP_Law1}, which says that the empirical spectral distribution of the matrix $\frac1N\mathbf G\mathbf G^T$, wherein $\mathbf G\in\mathbb R^{D\times N}$ has i.i.d entries sampled from $\mathcal N(0,1)$, converges almost surely to the deterministic distribution $F_{\gamma}(\lambda)$ as $D/N\longrightarrow\gamma\in[0,1)$.

Further, we need the following lemma on the inverse of the eigenvalues $\Lambda(\widehat{\boldsymbol\Sigma}_0)$, which says that the energy of $\Lambda^{-1}(\widehat{\boldsymbol\Sigma}_0)$ and $\Lambda^{-2}(\widehat{\boldsymbol\Sigma}_0)$ projected onto a random direction is almost surely deterministic in the limit. It is worth noticing that the results in Lemma \ref{lm:minus1moment} generalize the results on the expectations $\mathbb E[\sum_i{\lambda_i^{-1}}(\widehat{\boldsymbol\Sigma}_0)]$ and $\mathbb E[\sum_i{\lambda_i^{-2}}(\widehat{\boldsymbol\Sigma}_0)]$ in \cite{Hoyle_2011_pami}.
\begin{mylm}\label{lm:minus1moment} Suppose $\xi$ is a unit-length random vector uniformly distributed on the unit sphere $\mathbb S^{D-1}$ and it is independent of $\widehat{\boldsymbol\Sigma}_0$, then as $D/N\longrightarrow\gamma\in[0,1)$,
\begin{equation}\label{eq:minus1moment}
\begin{aligned}
    \xi^T\Lambda^{-1}(\widehat{\boldsymbol\Sigma}_0)\xi&\overset{a.s.}\longrightarrow\int \lambda^{-1} dF_{\gamma}(\lambda)
    =\frac{1}{1-\gamma},
\end{aligned}
\end{equation}
and
\begin{equation}\label{eq:minus2moment}
\begin{aligned}
    \xi^T\Lambda^{-2}(\widehat{\boldsymbol\Sigma}_0)\xi&\overset{a.s.}\longrightarrow\int \lambda^{-2} dF_{\gamma}(\lambda)=\frac1{(1-\gamma)^3}.
\end{aligned}
\end{equation}
\end{mylm}

\subsection{Properties of $\widehat{\mathbf S}_0$ }
We have the following lemma on the eigenvectors of $\widehat{\mathbf S}_0$.
\begin{mylm}\label{lm:eigenspace_S0}
Given the eigendecomposition $\widehat{\mathbf S}_0=\mathbf V\Lambda(\widehat{\mathbf S}_0)\mathbf V^T$, then as $D/N\longrightarrow\gamma\in[0,1)$,
\begin{equation}\label{eq:eigenspace_S0}
    \lim_{D/N\longrightarrow\gamma}\|\mathbf V_{1:c}^T\mathbf e_i\|^2 \ge\frac{\boldsymbol\lambda_i}{\boldsymbol\lambda_i+\gamma}, ~a.s., ~i=1,2,...,c,
\end{equation}
where $\boldsymbol\lambda_i$ is from the population discrimination power $\Delta(\boldsymbol\Sigma,\mathbf S|\mathbf W)=\sum_{i=1}^c\boldsymbol\lambda_i$.
\end{mylm}

Recalling Lemma \ref{lm:double_diagonal}, the population counterpart of $\widehat{\mathbf S}_0$ is actually the diagonal matrix $\boldsymbol\Lambda_0=\mathbf X^T\mathbf S\mathbf X$. Therefore, we expect the first $c$ eigenvectors $\mathbf V_{1:c}$ of $\widehat{\mathbf S}_0$ to be close to $\mathbf I_{1:c}=[\mathbf e_1,...,\mathbf e_c]$. Lemma \ref{lm:eigenspace_S0} shows that the performance of eigenvector estimation is determined by the $\boldsymbol\lambda_i$ and $\gamma$, and in particular, as $\gamma$ approaches $0$ the estimation becomes consistent.

\subsection{Proof of Theorem \ref{thm:generalization_bound}}
Now, we are ready to prove our main result Theorem \ref{thm:generalization_bound}, which is a conclusion out of the combination of Lemmas \ref{lm:gen_dis_pow}, \ref{lm:lowerbound_delta}, \ref{lm:minus1moment} and \ref{lm:eigenspace_S0}.

\begin{proof}
By Lemma \ref{lm:minus1moment}, we have
\begin{equation}\label{eq:lim_1}
    \lim_{D/N\longrightarrow\gamma}
    \frac{\xi^T\Lambda^{-1}(\widehat{\boldsymbol\Sigma}_0)\xi}{\sqrt{\xi^T\Lambda^{-2}(\widehat{\boldsymbol\Sigma}_0)\xi}}
    =\frac{\frac1{1-\gamma}}{\frac1{(1-\gamma)^{1.5}}}=\sqrt{1-\gamma}, \mbox{~a.s.}
\end{equation}
By Lemma \ref{lm:eigenspace_S0}, we have
\begin{equation}\label{eq:lim_2}
    \lim_{D/N\longrightarrow\gamma}\|\mathbf V_{1:c}^T\mathbf e_i\| \ge \sqrt{\boldsymbol\lambda_i/(\boldsymbol\lambda_i+\gamma)}, \mbox{~a.s.}
\end{equation}
Then the proof is completed by substituting (\ref{eq:lim_1}) and (\ref{eq:lim_2}) into Lemma \ref{lm:gen_dis_pow} and Lemma \ref{lm:lowerbound_delta}.
\end{proof}

\section{Empirical Evaluations}\label{sec:chp_gener_bound4}

\subsection{On the Bound of Generalization Discrimination Power}
According to Theorem \ref{thm:generalization_bound}, the generalization discrimination power of FLDA for dimension reduction can be factorized as  $\Delta(\boldsymbol\Sigma,\mathbf S|\widehat{\mathbf W}) =\sum_{i=1}^c\boldsymbol\delta_i\boldsymbol\lambda_i$, where $\boldsymbol\lambda_i$ measures the population discrimination power, and each component $\boldsymbol\delta_i\boldsymbol\lambda_i$ of the generalization discrimination power can be lower bounded by
\begin{equation}\label{eq:eta_cal}
    \boldsymbol\delta_i\boldsymbol\lambda_i\ge{\max}^2\big\{\cos(\arccos(\sqrt{{\boldsymbol\lambda_i}/(\boldsymbol\lambda_i + \gamma)}) + \arccos(\sqrt{1-\gamma})),0\big\}\boldsymbol\lambda_i.\notag
\end{equation}
We evaluate this result on both simulated and real datasets by comparing $\boldsymbol\delta_i\boldsymbol\lambda_i$ with the lower bound above.

\begin{figure}[thb]
\centering
\subfigure[Component $\boldsymbol\delta_1\boldsymbol\lambda_1$]{\includegraphics[width=0.495\columnwidth]{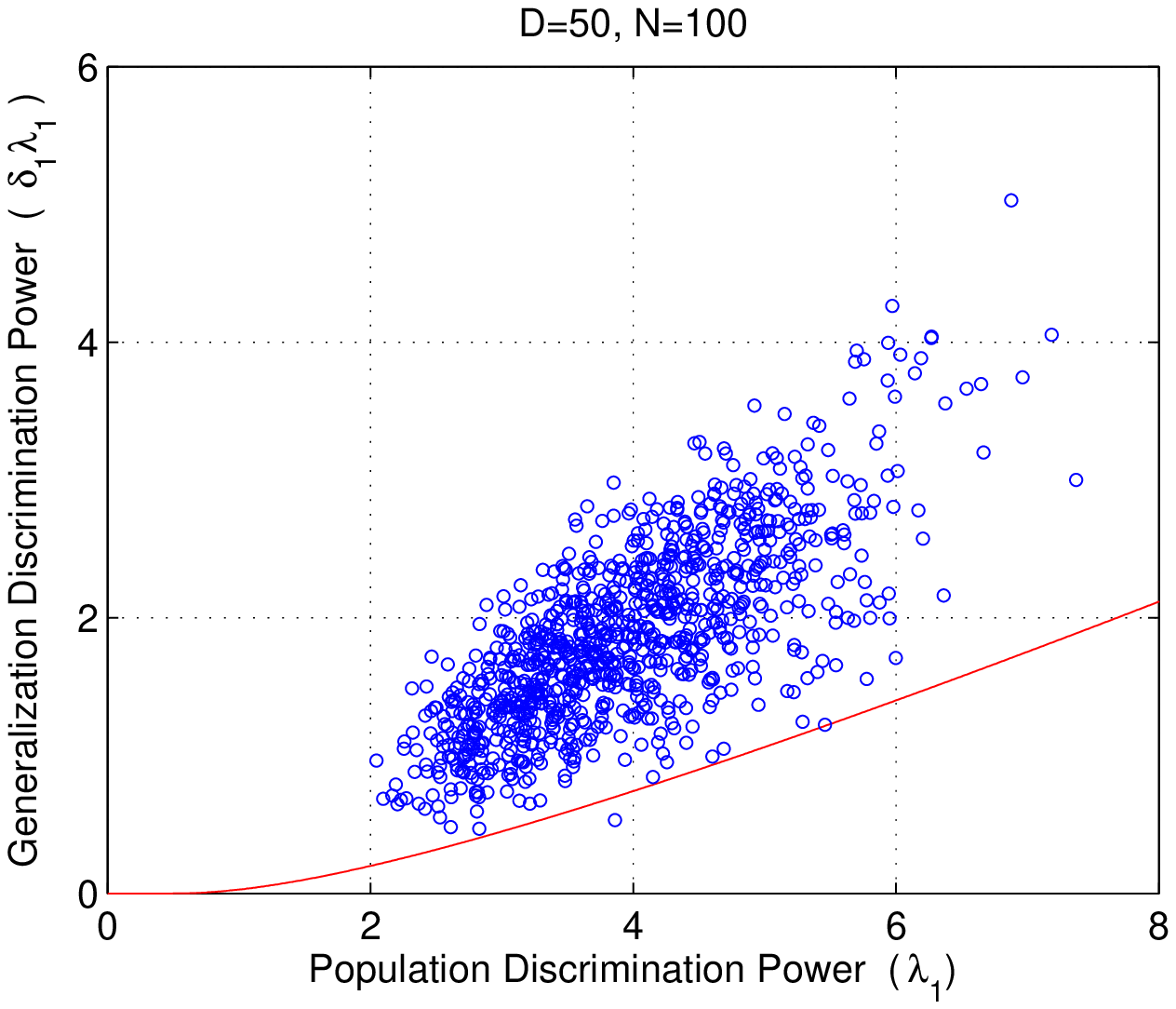}}
\subfigure[Component $\boldsymbol\delta_2\boldsymbol\lambda_2$]{\includegraphics[width=0.495\columnwidth]{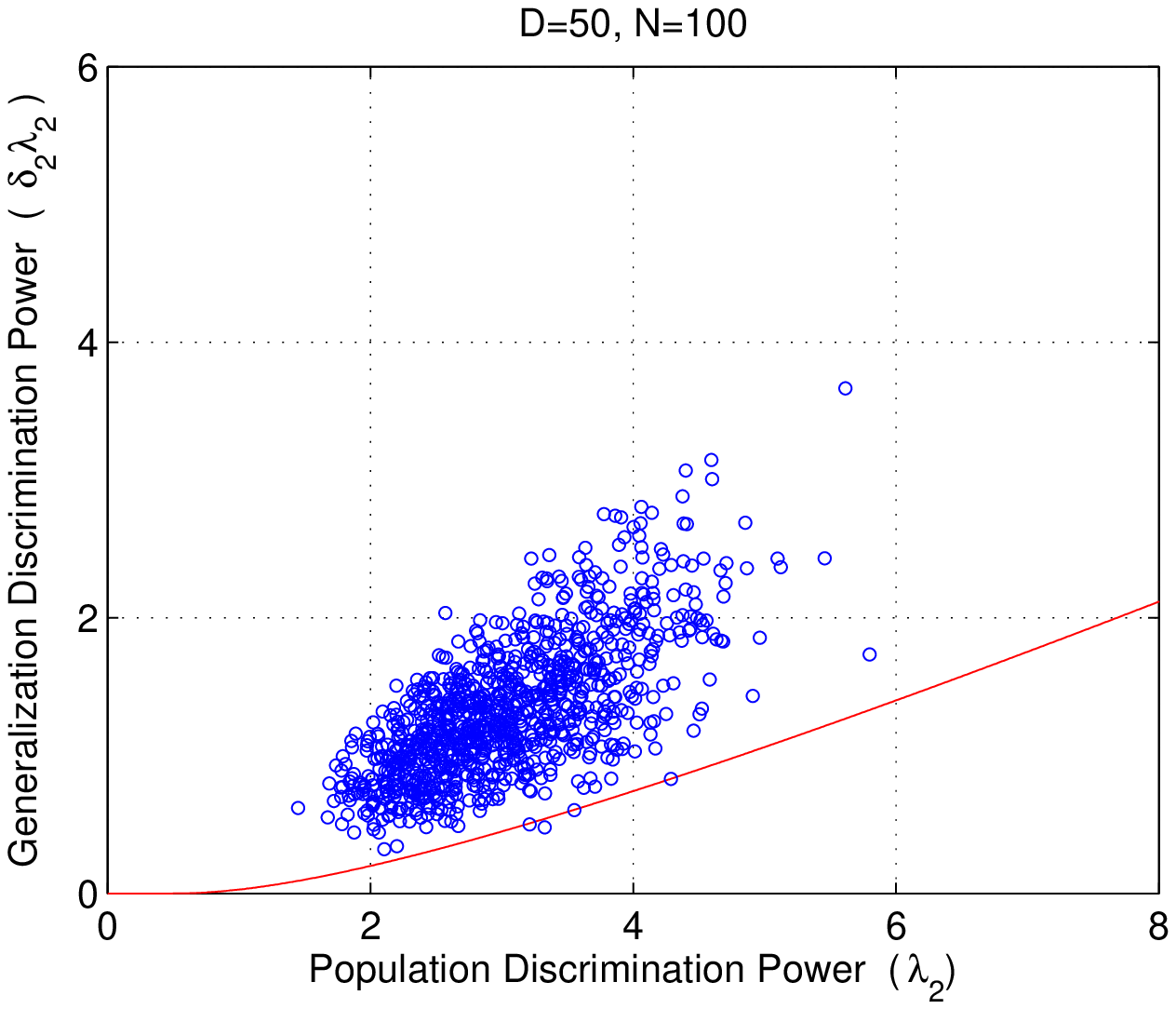}}\\
\subfigure[Component $\boldsymbol\delta_3\boldsymbol\lambda_3$]{\includegraphics[width=0.495\columnwidth]{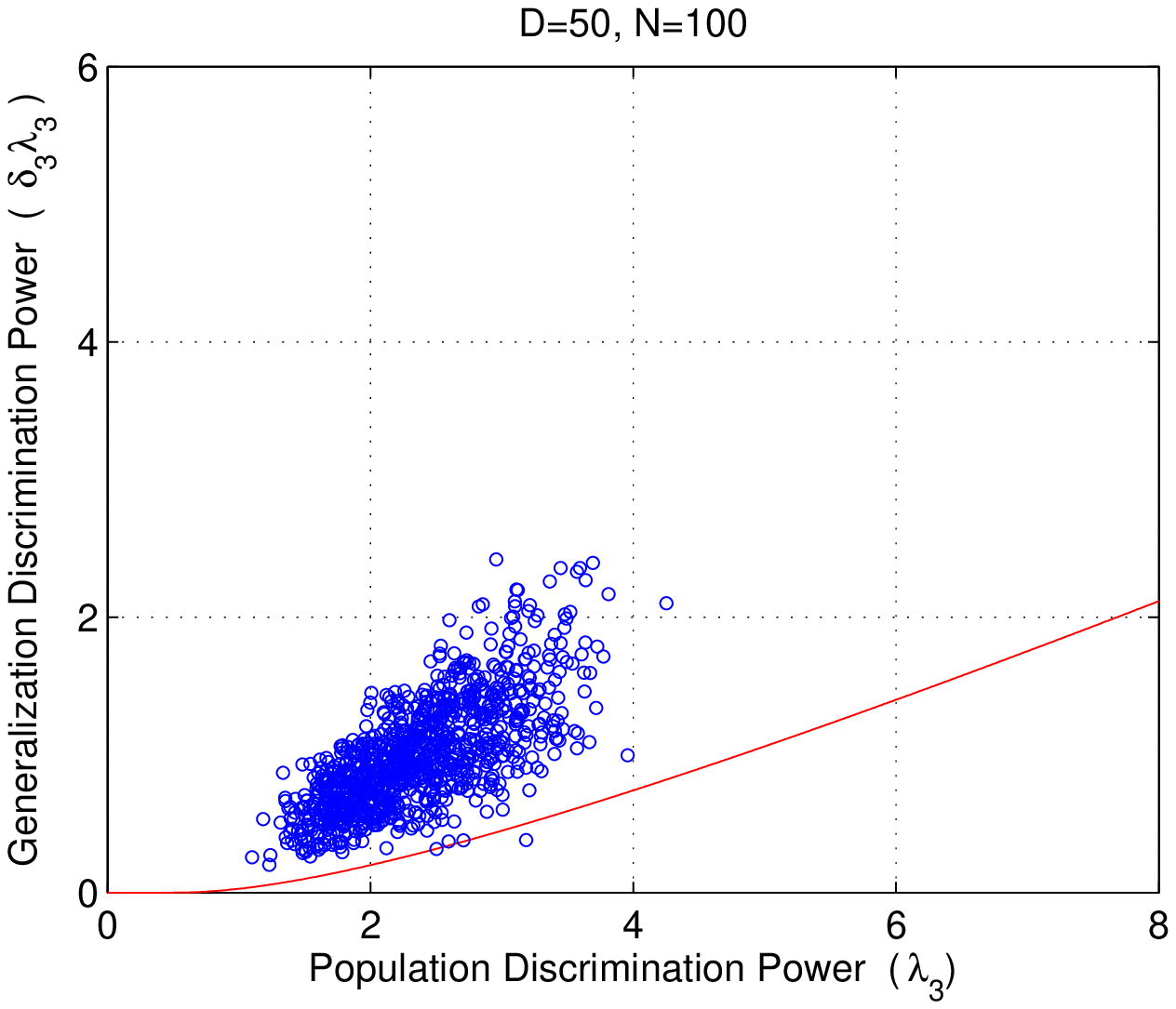}}
\subfigure[Component $\boldsymbol\delta_4\boldsymbol\lambda_4$]{\includegraphics[width=0.495\columnwidth]{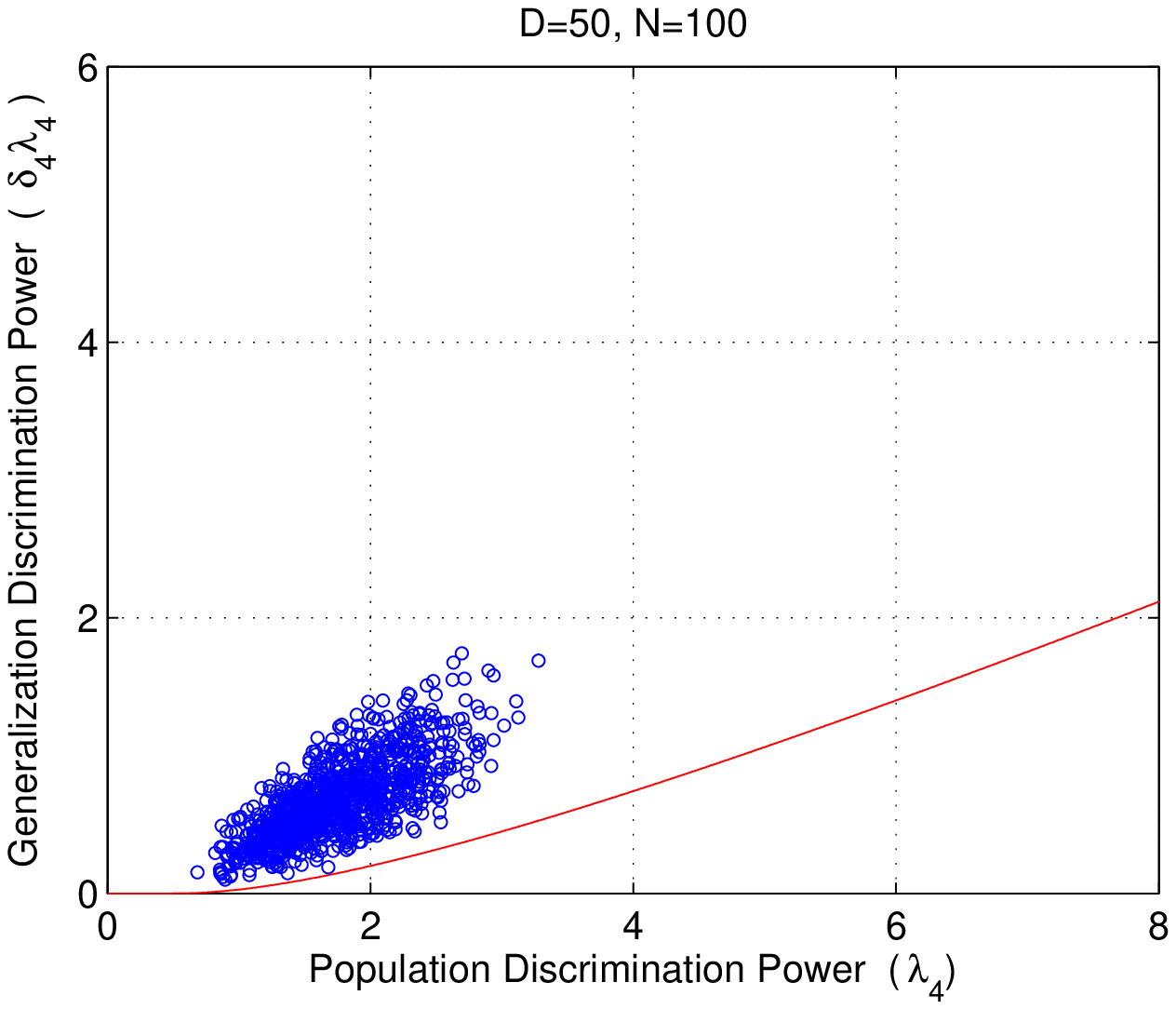}}\\
\caption{Evaluation of the Generalization Discrimination Power Bound with Simulated Data.}\label{fig:evaluaion_power_bound}
\end{figure}

For simulated data, we fix the ratio $\gamma=D/N=0.5$, with $D=50$ and $N=100$. Note the settings give moderate size problems; however, due to the asymptotic characteristic of the bound, which inherently fits to large size problem, the evaluation on moderate size problems is more critical. We generate 1,000 experiments, each having 5 classes with randomly generated population covariance $\boldsymbol\Sigma$ and class means $\boldsymbol\mu_i$, $i=1,...,5$. The population discrimination power $\boldsymbol\lambda_i$, $i=1,...,4$, are calculated via eigendecomposition of $\boldsymbol\Sigma^{-1}\mathbf S$, where $\mathbf S$ is the between-class scatter matrix. For the generalization discrimination power $\boldsymbol\delta_i\boldsymbol\lambda_i$, the factor $\boldsymbol\delta_i$ has a close form formulation as shown by Lemma \ref{lm:gen_dis_pow}, i.e.,
\begin{equation}\label{eq:delta_cal}\notag
\boldsymbol\delta_i=\|\mathcal R^T(\Lambda^{-1}(\widehat{\boldsymbol\Sigma}_0)\mathbf U^T\mathbf V_{1:c})\mathbf U^T\mathbf e_i\|^2,
\end{equation}
where $\Lambda(\widehat{\boldsymbol\Sigma}_0)$ and $\mathbf U$ are the eigensystems of $\widehat{\boldsymbol\Sigma}_0$ and $\mathbf V_{1:c}$ are the first $c$ eigenvectors of $\widehat{\mathbf S}_0$, with $\widehat{\boldsymbol \Sigma}_0 = \mathbf X^{T}\widehat{\boldsymbol \Sigma}\mathbf X$ and $\widehat{\mathbf S}_0 =  \mathbf X^{T}\widehat{\mathbf S}\mathbf X$ being the normalized sample covariance and between-class scatter matrix and $\mathbf X$ simultaneously diagonalizing $\boldsymbol\Sigma$ and $\mathbf S$. Since a larger discrimination power means a better separation between classes, we expect that on most of the experiments the generalization discrimination power of FLDA can be bounded from the lower side by the generalization bound. Indeed, as shown by Figure \ref{fig:evaluaion_power_bound}, the bound holds with an overwhelming probability in the empirical sense (i.e., on more than 990 out of the 1,000 experiments).

\begin{figure}[thb]
\centering
\subfigure[ImageSeg]{\includegraphics[width=0.495\columnwidth]{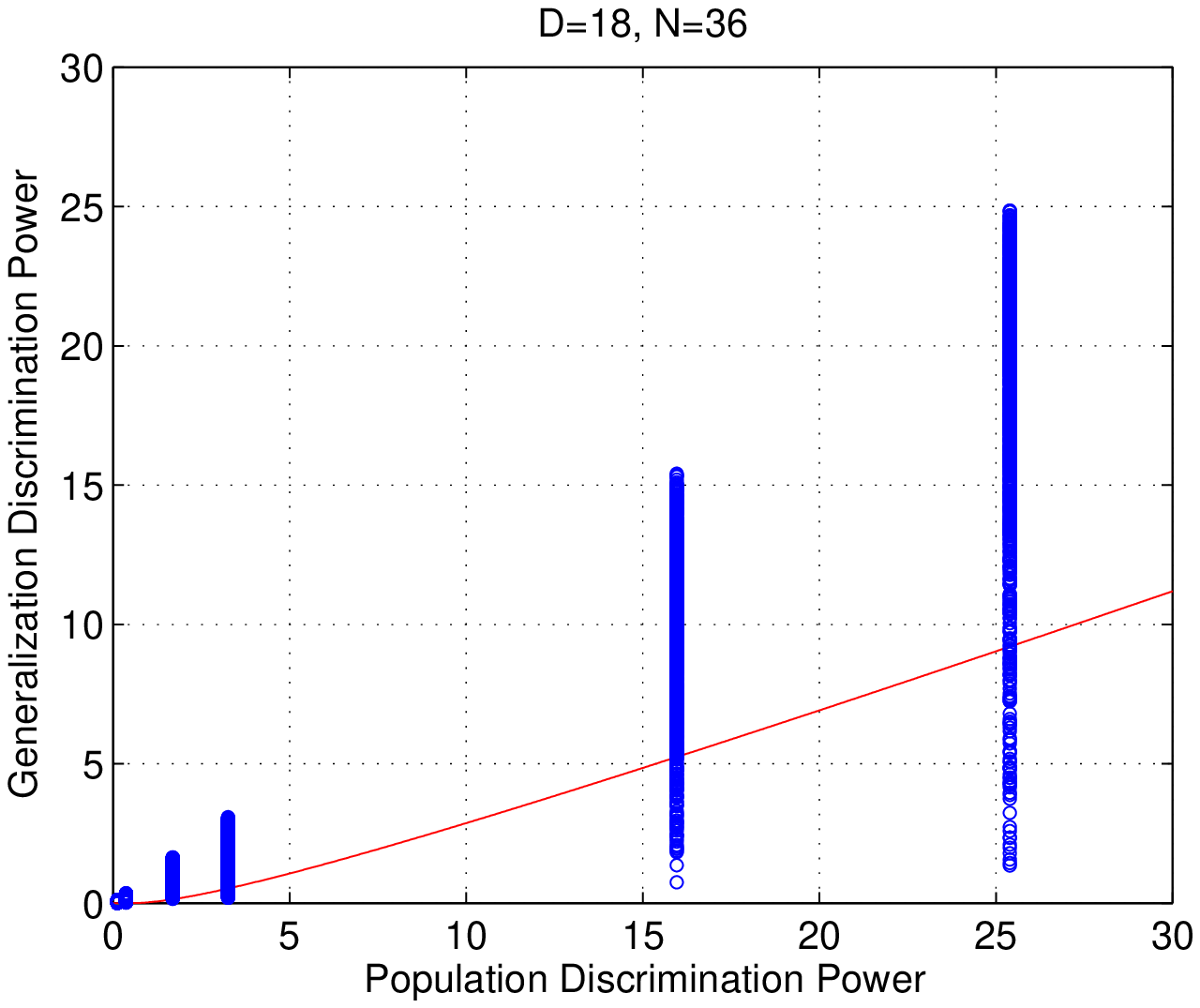}}
\subfigure[LandSat]{\includegraphics[width=0.495\columnwidth]{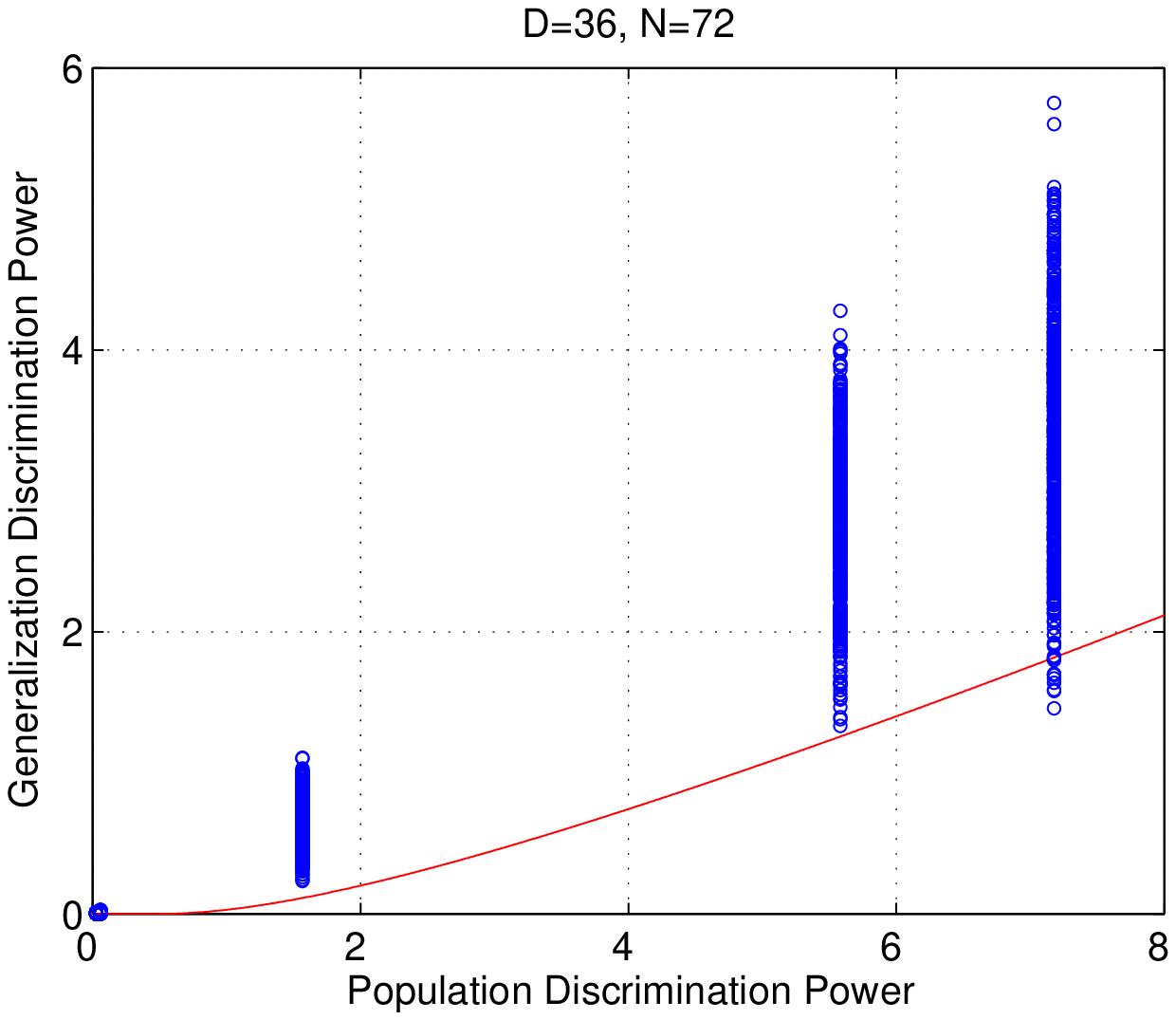}}\\
\subfigure[Optdigits]{\includegraphics[width=0.495\columnwidth]{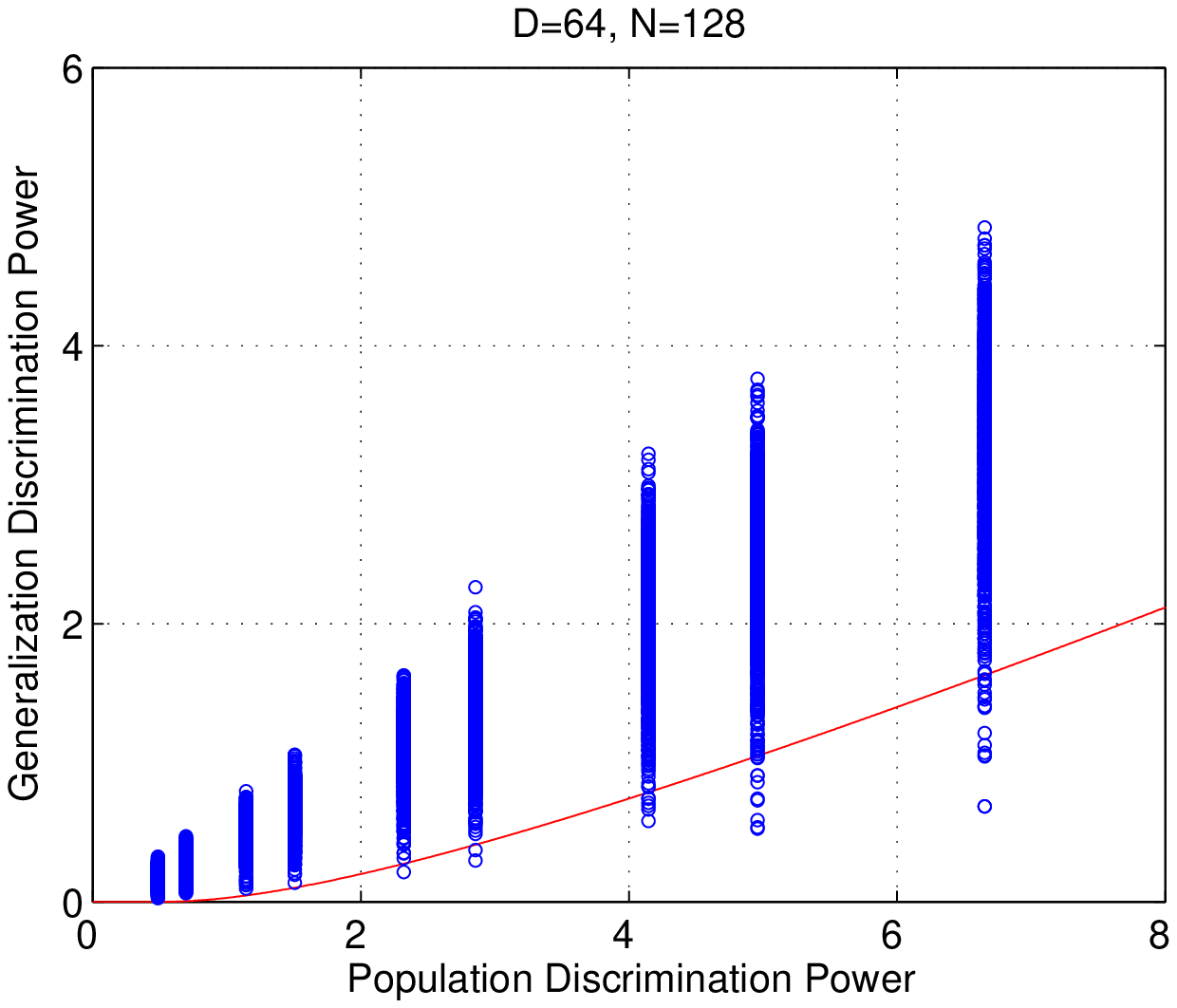}}
\subfigure[USPS]{\includegraphics[width=0.495\columnwidth]{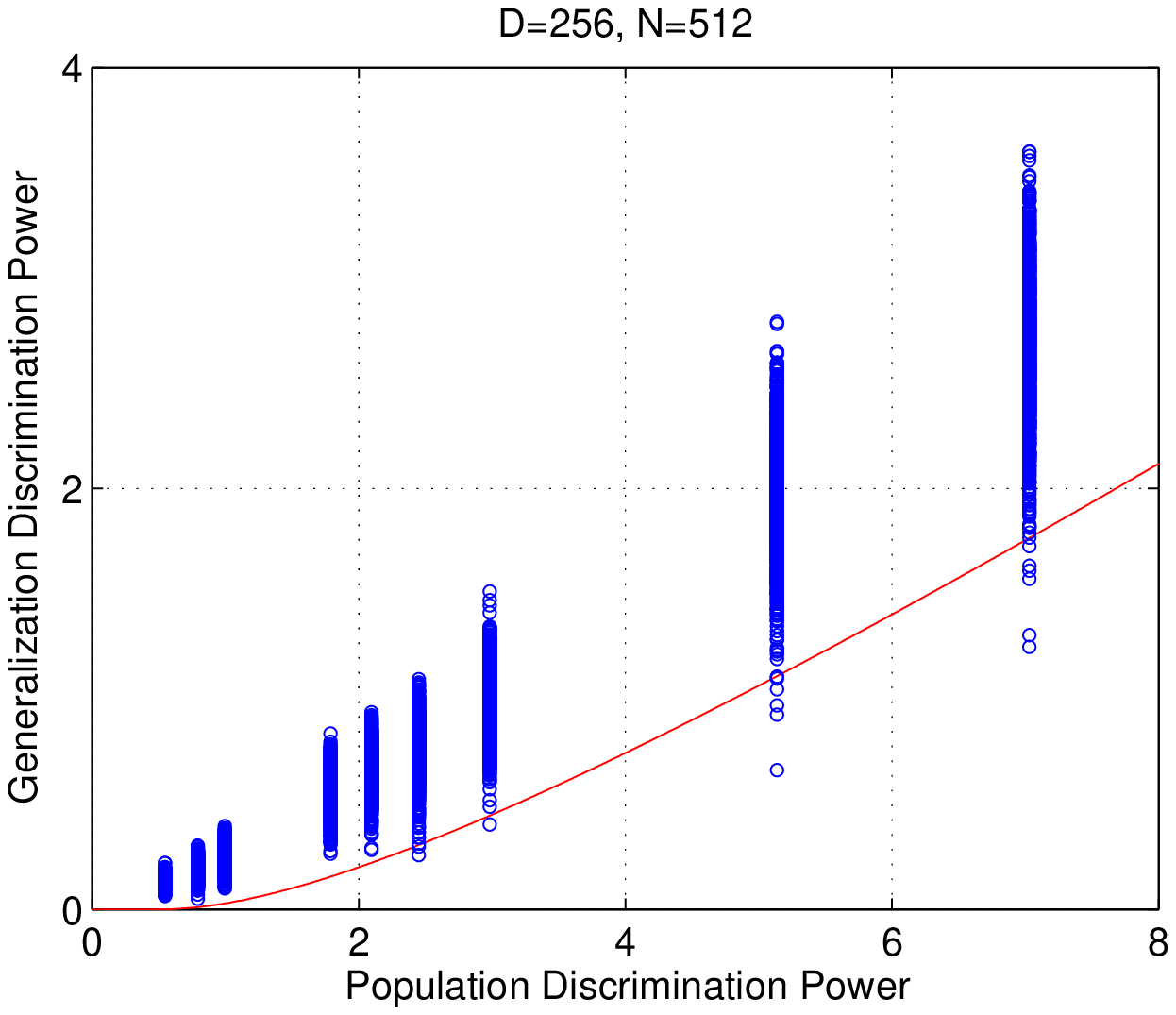}}\\
\caption{Evaluation of the Generalization Discrimination Power Bound with Real Data.}\label{fig:evaluaion_power_bound_real_data}
\end{figure}
We further evaluate the bound of generalization discrimination power on four benchmark datesets from the UCI machine learning repository \cite{Blake+Merz:1998}: 1) the image segmentation (ImageSeg) dataset \footnote{The original dataset has 19 features; however the 3rd feature is a constant for all examples, and therefore is discarded in the experiments.}, which contains 7 classes and in total 2,310 examples from $\mathbb R^{18}$; 2) the Landsat dataset, which constants 6 classes and in total 6,435 examples from $\mathbb R^{36}$; 3) the optical recognition of handwritten digits (Optdigits) dataset, which contains 10 classes and in total 5,620 examples from $\mathbb R^{60}$; and 4) the USPS handwritten digits dataset, which contains 10 classes and in total 9,298 examples from $\mathbb R^{256}$. Note that for real dataset, the population parameters $\boldsymbol\Sigma$ and $\mathbf S$ are unknown. Thus, we use the entire dataset to get their estimates and treat them as population parameters. Again, we fix the ratio $\gamma=D/N=0.5$, i.e., we randomly select examples twice of the dimensionality as the training data. The generalization discrimination powers over 1,000 random experiments are shown in Figure \ref{fig:evaluaion_power_bound_real_data}. On the panel for each dataset, the columns of the scatters correspond to different components of the generalization discrimination power $\boldsymbol\delta_i\boldsymbol\lambda_i$, and the horizontal axis location of each column equals the population discrimination power $\boldsymbol\lambda_i$ (the column number is class number minus 1). On three out of the four datasets, including LandSat, Optdigits and USPS, the generalization discrimination power is properly bounded by the lower bound, with a high probability in the empirical sense. On the ImageSeg dataset, the bound does not hold with high probability as on the other three datasets. The major reason is that the size of the problem is considerably small, with $D=18$ and $N=36$, while the bound favors large or moderate size problems.

\subsection{On the Bound of Generalization Errors}

According to Corollary \ref{coly:generalization_bound}, suppose the Bayes error of a binary classification problem is $P_{Bayes}$, then the generalization error $P$ of FLDA can be boudned by
\begin{align}
  P\le \Phi(\varrho\Phi^{-1}(P_{Bayes}))\notag,
\end{align}
where $\Phi(\cdot)$ is the CDF of the standard Gaussian distribution and
\begin{equation}\label{eq:varrho}\notag
    \varrho={\max}\left\{\cos\left(\arccos\left(\sqrt{\frac{(\Phi^{-1}(P_{Bayes}))^2}{((\Phi^{-1}(P_{Bayes}))^2 + \gamma}}\right) + \arccos(\sqrt{1-\gamma})\right),0\right\}.
\end{equation}
\begin{figure}
\centering
\includegraphics[width=0.495\columnwidth]{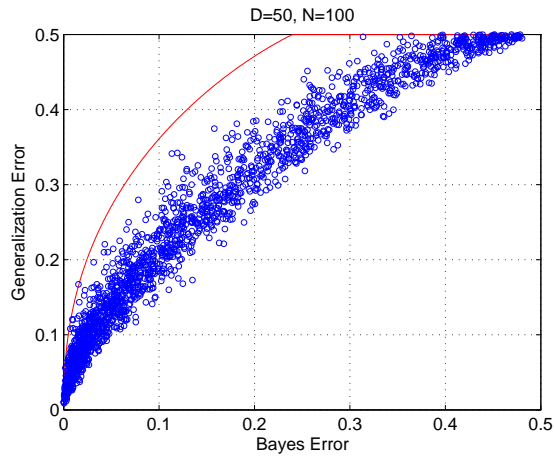}
\caption{Evaluation of the Generalization Error Bound with Simulated Data.}\label{fig:generalization_error_bound_lda}
\end{figure}

To evaluate this result, we perform binary classification with FLDA on 1,000 experiments, with randomly generated covariance matrix and class means. The same as in previous simulation, we fix the ratio $\gamma=D/N=0.5$, with $D=50$ and $N=100$. Figure \ref{fig:generalization_error_bound_lda} shows the result, where the generalization error of FLDA is properly bounded by the upper bound.

\begin{figure}
\centering
\subfigure[ImageSeg]{\includegraphics[width=0.495\columnwidth]{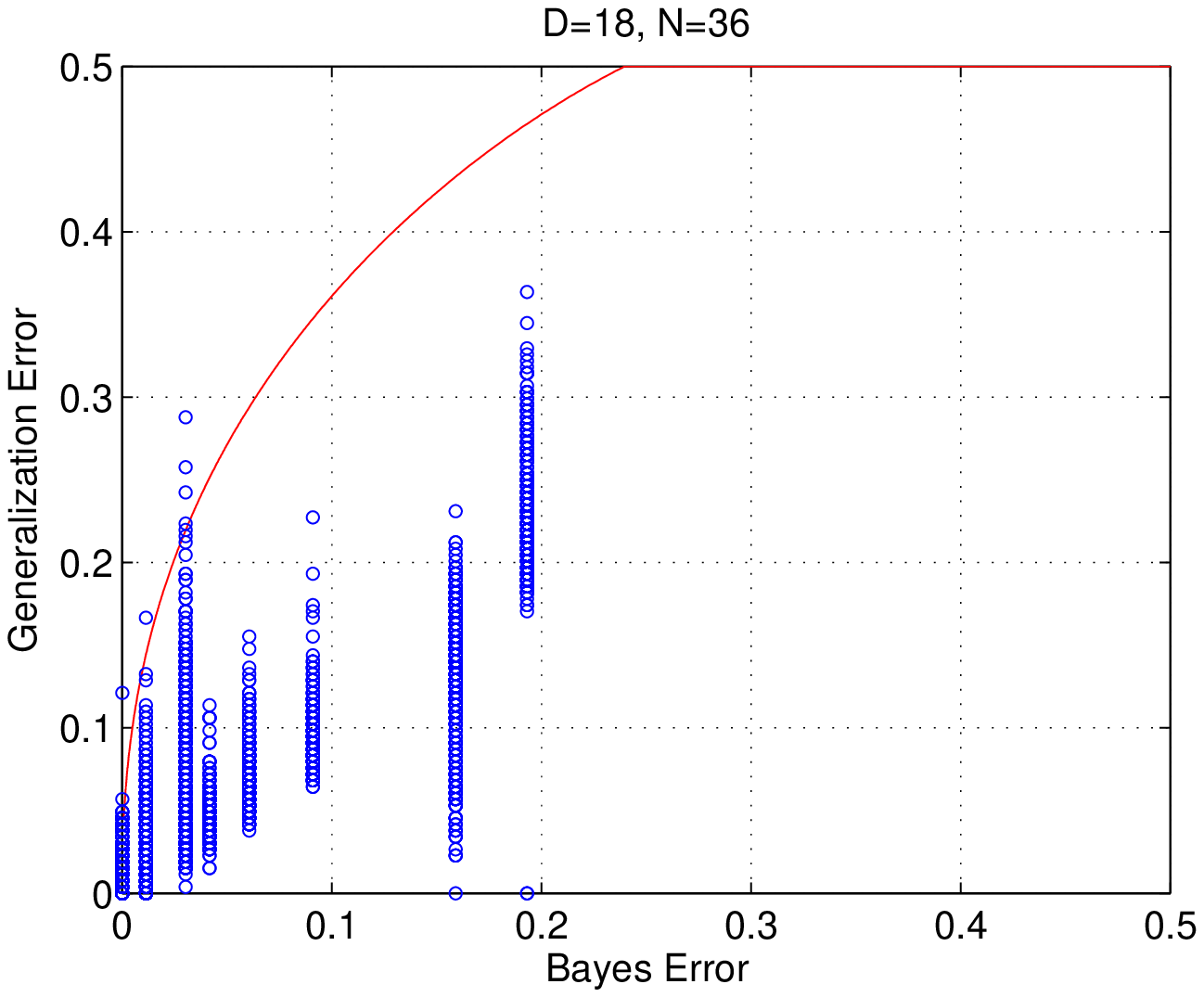}}
\subfigure[LandSat]{\includegraphics[width=0.495\columnwidth]{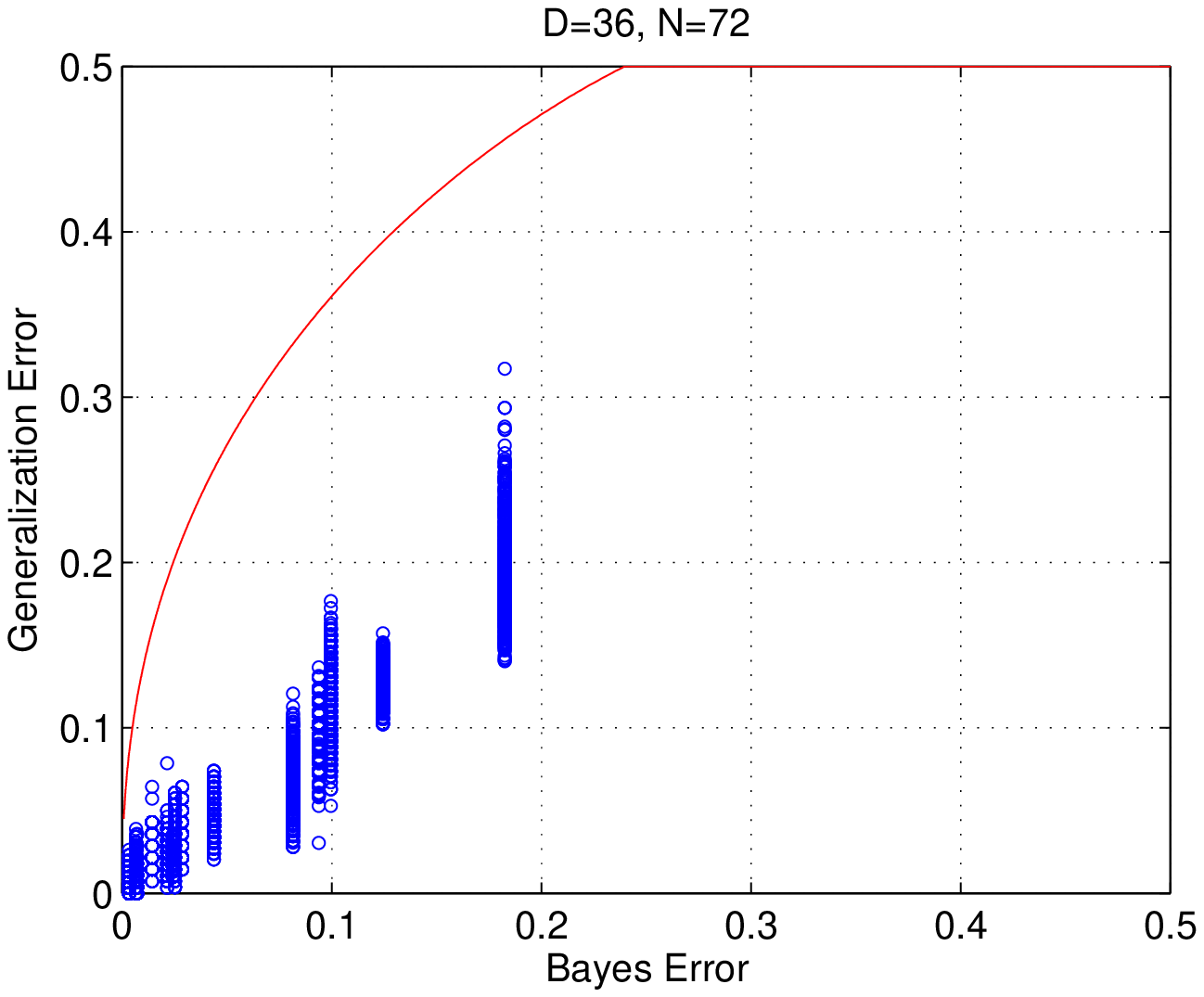}}\\
\subfigure[Optdigits]{\includegraphics[width=0.495\columnwidth]{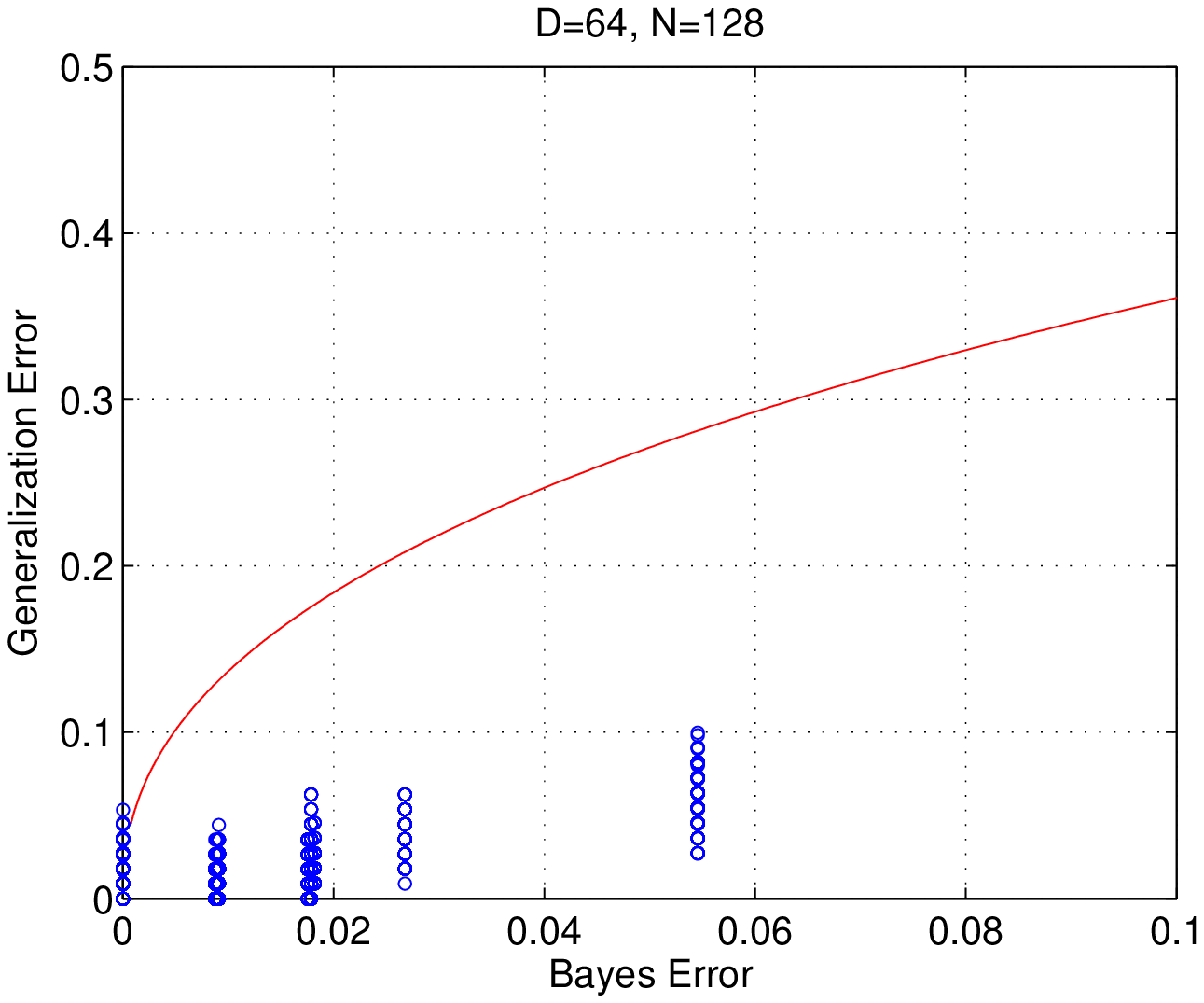}}
\subfigure[USPS]{\includegraphics[width=0.495\columnwidth]{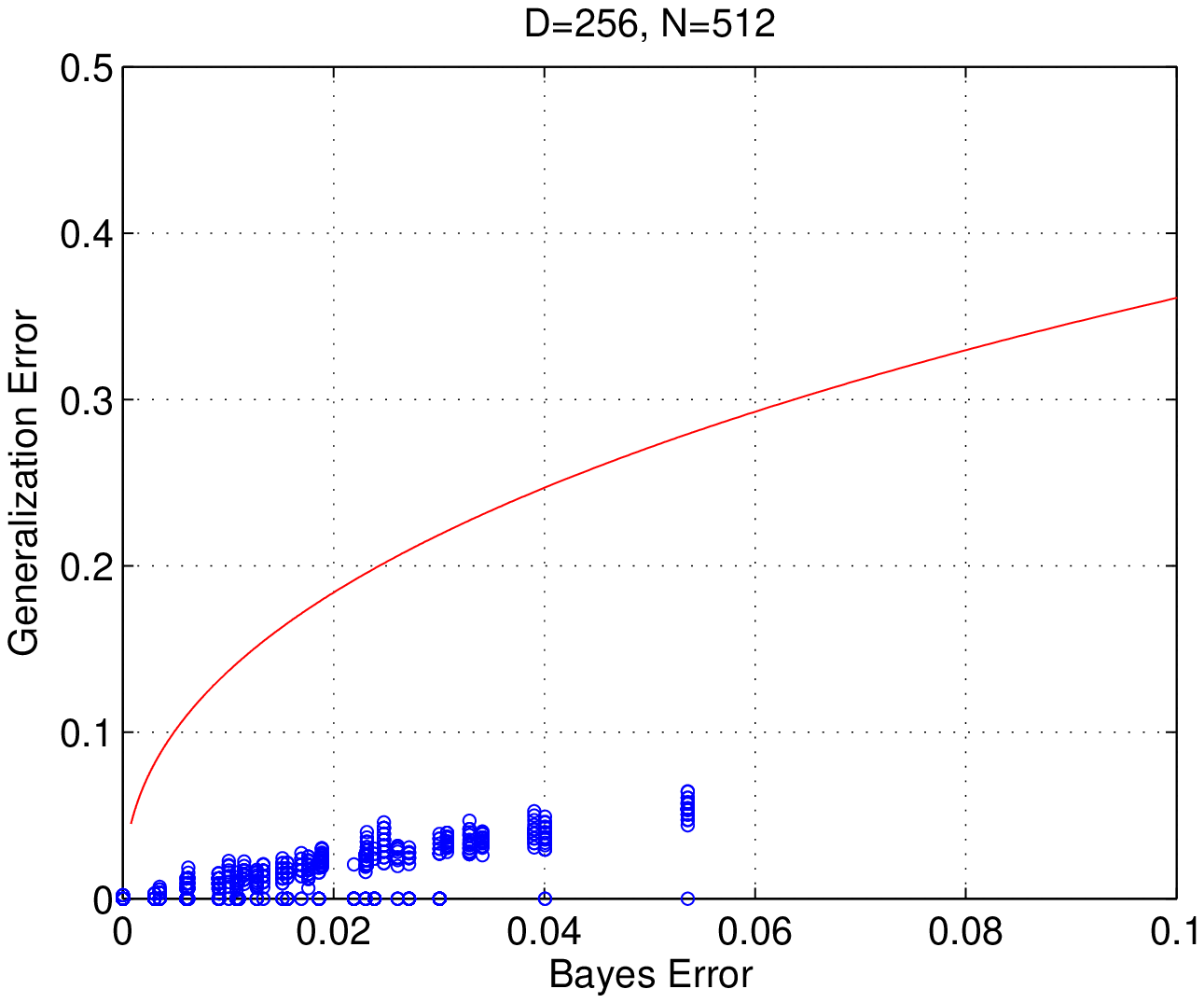}}\\
\caption{Evaluation of the Generalization Error Bound with Real Data.}\label{fig:evaluation_error_bound_real}
\end{figure}

In addition, we run experiments on the previous four real datasets to evaluate the generalization error bound. We randomly select class pairs from each dataset to perform binary classification. We hold out 10\% data as the evaluation set, which is used to estimate the ``Bayes'' error and generalization error. The ``Bayes'' classifier is obtained by training FLDA on the rest 90\% data, and the empirical classifier is trained with a subset of the rest data, such that $N=2D$, namely fixing the ratio $\gamma=D/N=0.5$. On each dataset, 1,000 random experiments are performed, with the result shown in Figure \ref{fig:evaluation_error_bound_real}. Similar to the result in Figure \ref{fig:evaluaion_power_bound_real_data}, on three out of the four datasets, the generalization error can be bounded by the upper bound, while the bound does not dominate all the experiment on the ImageSeg dataset due to the small size of the problem.

\section{Proofs of Lemmas and Corollary}\label{sec:chp_gener_bound5}
This section provides detailed proofs of Lemmas in Section III and Corollary \ref{coly:generalization_bound} in Section II.

\subsection{Proof of Lemma \ref{lm:double_diagonal}}
It is a direct result of the simultaneous diagonalization theorem for a pair of semidefinite matrices \cite{FukunagaBook}.
\subsection{Proof of Lemma \ref{lm:gen_dis_pow}}
The proof is divided into two steps.

\noindent
i) Since $\mathbf X$ in Lemma \ref{lm:double_diagonal} is nonsingular, there exists some $\mathbf Q\in\mathbb R^{D\times c}$ such that $\widehat{\mathbf W} = \mathbf X\mathbf Q$. Then,

\begin{equation}
\begin{aligned}
    \Delta(\boldsymbol\Sigma,\mathbf S|\widehat{\mathbf W})
    &=\mbox{Tr}((\widehat{\mathbf W}^{T}\boldsymbol\Sigma\widehat{\mathbf W})^{-1}\widehat{\mathbf W}^{T}\mathbf S\widehat{\mathbf W})\\
    &=\mbox{Tr}((\mathbf Q^T\mathbf X^{T}\boldsymbol\Sigma \mathbf X\mathbf Q)^{-1}\mathbf Q^T\mathbf X^{T}\mathbf S\mathbf X\mathbf Q)\\
    &=\mbox{Tr}((\mathbf Q^T\mathbf Q)^{-1}\mathbf Q^T\mathbf X^{T}\boldsymbol\Lambda\mathbf Q)\\
    &=\mbox{Tr}((\mathbf Q^T\mathbf Q)^{-1}\mathbf Q_1^T\boldsymbol\Lambda_1\mathbf Q_1)\\
    &=\mbox{Tr}(\mathbf Q_1(\mathbf Q^T\mathbf Q)^{-1}\mathbf Q_1^T\boldsymbol\Lambda_1)\\
    &=\sum_{i=1}^c\boldsymbol\delta_i\boldsymbol\lambda_i,\\
\end{aligned}
\end{equation}
where $\mathbf Q_1$ contains the first $c$ rows of $\mathbf Q$ and $\boldsymbol\Lambda_1$ is the upper-left $c\times c$ submatrix of $\boldsymbol\Lambda$, and clearly,
\begin{equation}
    \boldsymbol\delta_i = \{\mathbf Q_1(\mathbf Q^T\mathbf Q)^{-1}\mathbf Q_1^T\}_{ii}.
\end{equation}

\noindent
ii) In FLDA, $\widehat{\mathbf W}$ are the eigenvectors of $\widehat{\boldsymbol\Sigma}^{-1}\widehat{\mathbf S}$, and we can restrict the scale of $\widehat{\mathbf W}$ such that
\begin{equation} \label{eq:diag_empirical}
    \widehat{\mathbf W}^{T}\widehat{\boldsymbol\Sigma}\widehat{\mathbf W}=\mathbf I_c \mbox{~and~} \widehat{\mathbf W}^{T}\widehat{\mathbf S}\widehat{\mathbf W}=\widehat{\boldsymbol\Lambda}_1,
\end{equation}
where $\widehat{\boldsymbol\Lambda}_1$ is some $c\times c$ diagonal matrix. Substituting $\widehat{\mathbf W} = \mathbf X\mathbf Q$ into (\ref{eq:diag_empirical}) and recalling  $\widehat{\boldsymbol \Sigma}_0 = \mathbf X^{T}\widehat{\boldsymbol \Sigma}\mathbf X$ and $\widehat{\mathbf S}_0 =  \mathbf X^{T}\widehat{\mathbf S}\mathbf X$, we get
\begin{equation}\label{eq:define_Q}
    \mathbf Q^T\widehat{\boldsymbol \Sigma}_0\mathbf Q=\mathbf I_c \mbox{~and~} \mathbf Q^T\widehat{\mathbf S}_0\mathbf Q=\widehat{\boldsymbol\Lambda}_1.
\end{equation}

Given the eigendecomposition $\widehat{\boldsymbol\Sigma}_0=\mathbf U\Lambda(\widehat{\boldsymbol\Sigma}_0)\mathbf U^T$, we have from the first equation in (\ref{eq:define_Q}) that there must exist some orthogonal matrix $\mathbf O\in\mathbb R^{D\times c}$, $\mathbf O^T\mathbf O=\mathbf I_c$, such that
\begin{equation}\label{eq:cal_Q}
\mathbf Q=\mathbf U\Lambda^{-\frac12}(\widehat{\boldsymbol\Sigma}_0)\mathbf O.
\end{equation}
Further, given the eigendecomposition $\widehat{\mathbf S}_0=\mathbf V^T\Lambda(\widehat{\mathbf S}_0)\mathbf V$, we get from the second equation in (\ref{eq:define_Q}) that
\begin{equation}\label{eq:cal_O}
\mathbf O^T\Lambda^{-\frac12}(\widehat{\boldsymbol\Sigma}_0)\mathbf U^T \mathbf V\Lambda(\widehat{\mathbf S}_0)\mathbf V^T\mathbf U\Lambda^{-\frac12}(\widehat{\boldsymbol\Sigma}_0)\mathbf O=\widehat{\boldsymbol\Lambda}_1.
\end{equation}
In addition, since $\widehat{\mathbf S}_0$ has rank $c$, we can rewrite (\ref{eq:cal_O}) as
\begin{equation}\label{eq:tmp10}
\mathbf O^T\Lambda^{-\frac12}(\widehat{\boldsymbol\Sigma}_0)\mathbf U^T \mathbf V_{1:c}\Lambda_1^{\frac12}(\widehat{\mathbf S}_0)\Lambda_1^{\frac12}(\widehat{\mathbf S}_0)\mathbf V_{1:c}^T\mathbf U\Lambda^{-\frac12}(\widehat{\boldsymbol\Sigma}_0)\mathbf O=\widehat{\boldsymbol\Lambda}_1,
\end{equation}
where $\Lambda_1(\widehat{\boldsymbol\Sigma}_0)$ is the upper-left $c\times c$ submatrix of $\Lambda(\widehat{\boldsymbol\Sigma}_0)$. (\ref{eq:tmp10}) implies the columns of $\mathbf O$ must be the left singular vectors of $\Lambda^{-\frac12}(\widehat{\boldsymbol\Sigma}_0)\mathbf U^T \mathbf V_{1:c}\Lambda_1^{\frac12}(\widehat{\mathbf S}_0)$. Thus, $\mathbf O$ spans the range space of $\Lambda^{-\frac12}(\widehat{\boldsymbol\Sigma}_0)\mathbf U^T \mathbf V_{1:c}\Lambda_1^{\frac12}(\widehat{\mathbf S}_0)$ and therefore the range space of $\Lambda^{-\frac12}(\widehat{\boldsymbol\Sigma}_0)\mathbf U^T \mathbf V_{1:c}$. Then, there must exist some matrix $\mathbf A\in\mathbb R^{c\times c}$ such that $\Lambda^{-\frac12}(\widehat{\boldsymbol\Sigma}_0)\mathbf U^T \mathbf V_{1:c}=\mathbf O\mathbf A$, and thus
\begin{equation}\label{eq:tmp_O}
 \mathbf O=\Lambda^{-\frac12}(\widehat{\boldsymbol\Sigma}_0)\mathbf U^T \mathbf V_{1:c}\mathbf A^{-1},
\end{equation}
where the nonsingularity of $\mathbf A$ is implied by the nonsingularity of $\Lambda^{-\frac12}(\widehat{\boldsymbol\Sigma}_0)\mathbf U^T$.

By (\ref{eq:cal_Q}) and (\ref{eq:tmp_O}), we have
\begin{equation}\label{eq:Q_useful}
    \mathbf Q=\mathbf U\Lambda^{-1}(\widehat{\boldsymbol\Sigma}_0)\mathbf U^T \mathbf V_{1:c}\mathbf A,
\end{equation}
and
\begin{equation}
    \mathbf Q_1=\mathbf I_{1:c}^T\mathbf U\Lambda^{-1}(\widehat{\boldsymbol\Sigma}_0)\mathbf U^T \mathbf V_{1:c}\mathbf A.
\end{equation}
Therefore,
\begin{equation}\label{eq:Q_1QQQ_1}
\begin{aligned}
    \{\mathbf Q_1&(\mathbf Q^T\mathbf Q)^{-1}\mathbf Q_1\}_{ii} =
   \mathbf e_i^T\mathbf U\Lambda^{-1}(\widehat{\boldsymbol\Sigma}_0)\mathbf U^T \mathbf V_{1:c}(\mathbf V_{1:c}^T\mathbf U\Lambda^{-2}(\widehat{\boldsymbol\Sigma}_0)\mathbf U^T\mathbf V_{1:c})^{-1}\mathbf V_{1:c}^T\mathbf U\Lambda^{-1}(\widehat{\boldsymbol\Sigma}_0)\mathbf U^T\mathbf e_i.\\
\end{aligned}
\end{equation}
Letting $\mathbf R=\mathcal R(\Lambda^{-1}(\widehat{\boldsymbol\Sigma}_0)\mathbf U^T\mathbf V_{1:c})$, then
\begin{equation}
    \mathbf R\mathbf R^T = \Lambda^{-1}(\widehat{\boldsymbol\Sigma}_0)\mathbf U^T \mathbf V_{1:c}(\mathbf V_{1:c}^T\mathbf U\Lambda^{-2}(\widehat{\boldsymbol\Sigma}_0)\mathbf U^T\mathbf V_{1:c})^{-1}\mathbf V_{1:c}^T\mathbf U\Lambda^{-1}(\widehat{\boldsymbol\Sigma}_0),
\end{equation}
which together with (\ref{eq:Q_1QQQ_1}) gives
\begin{equation}
\begin{aligned}
    \{\mathbf Q_1(\mathbf Q_{\ell}^T\mathbf Q_{\ell})^{-1}\mathbf Q_1\}_{ii} &= \mathbf e_i^T\mathbf U\mathbf R\mathbf R^T \mathbf U^T\mathbf e_i
    =\|\mathbf R^T\mathbf U^T\mathbf e_i\|^2\\
    &=\|\mathcal R^T(\Lambda^{-1}(\widehat{\boldsymbol\Sigma}_0)\mathbf U^T\mathbf V_{1:c})\mathbf U^T\mathbf e_i\|^2.
    \end{aligned}
\end{equation}
This completes the proof.

\subsection{Proof of Lemma \ref{lm:lowerbound_delta}}
Recall Lemma \ref{lm:gen_dis_pow} that $\boldsymbol\delta_i=\|\mathcal R^T(\Lambda^{-1}(\widehat{\boldsymbol\Sigma}_0)\mathbf U^T\mathbf V_{1:c})\mathbf U^T\mathbf e_i\|^2$. Denote by $\measuredangle(\mathbf U^T\mathbf e_i,\mathcal R(\Lambda^{-1}(\widehat{\boldsymbol\Sigma}_0)\mathbf U^T\mathbf V_{1:c}))$ the angle between vector $\mathbf U^T\mathbf e_i$ and subspace $\mathcal R^T(\Lambda^{-1}(\widehat{\boldsymbol\Sigma}_0)\mathbf U^T\mathbf V_{1:c})$, we have
\begin{equation}
   \boldsymbol\delta_i
   =\cos^2(\measuredangle(\mathbf U^T\mathbf e_i,\mathcal R(\Lambda^{-1}(\widehat{\boldsymbol\Sigma}_0)\mathbf U^T\mathbf V_{1:c}))).
\end{equation}

Two basic facts that hold for arbitrary vector $\mathbf a_1$, $\mathbf a_2$ and subspace $\mathbf A$ are
\begin{equation}\label{eq:angle_trangle_1}
    \measuredangle(\mathbf a_1,\mathbf A)\le \measuredangle(\mathbf a_1,\mathbf a_2) + \measuredangle(\mathbf a_2,\mathbf A).
\end{equation}
and
\begin{equation}\label{eq:angle_trangle_2}
    \measuredangle(\mathbf a_1,\mathbf A)\le \measuredangle(\mathbf a_1,\mathbf a), \mbox{~if~} \mathbf a\in\mathbf A.
\end{equation}

Then, by using (\ref{eq:angle_trangle_1}) and (\ref{eq:angle_trangle_2}), we get
\begin{equation}
\begin{aligned}    &\measuredangle(\mathbf U^T\mathbf e_i,\mathcal R(\Lambda^{-1}(\widehat{\boldsymbol\Sigma}_0)\mathbf U^T\mathbf V_i))\\
\le& \measuredangle(\mathbf U^T\mathbf e_i,\mathbf U^T\mathbf V_{1:c}\mathbf V_{1:c}^T\mathbf e_i)+\measuredangle(\mathbf U^T\mathbf V_{1:c}\mathbf V_{1:c}^T\mathbf e_i,\mathcal R(\Lambda^{-1}(\widehat{\boldsymbol\Sigma}_0)\mathbf U^T\mathbf V_{1:c}))\\
\le& \measuredangle(\mathbf U^T\mathbf e_i,\mathbf U^T\mathbf V_{1:c}\mathbf V_{1:c}^T\mathbf e_i)+\measuredangle(\mathbf U^T\mathbf V_{1:c}\mathbf V_{1:c}^T\mathbf e_i,\Lambda^{-1}(\widehat{\boldsymbol\Sigma}_0)\mathbf U^T\mathbf V_{1:c}\mathbf V_{1:c}^T\mathbf e_i)\\
=&\theta_1 + \theta_2.
\end{aligned}
\end{equation}

Denoting $\theta=\theta_1 + \theta_2$, since $\cos(x)$ is positive and decreasing on $[0,\pi/2]$, $x^2$ is increasing on $[0,1]$, and $\boldsymbol\delta_i$ is nonnegative, we have
\begin{equation}
\begin{aligned}
    \boldsymbol\delta_i&\ge\left\{\begin{array}{ll}
    \cos^2(\theta),&\theta\le\frac\pi2\\
    0,&\mbox{else}
    \end{array}\right.\\
    &={\max}^2\{\cos(\theta),0\}.
\end{aligned}
\end{equation}

It remains to calculate $\theta_1$ and $\theta_2$. For $\theta_1$, We have
\begin{equation}
\begin{aligned}
    \cos^2(\theta_1) = \frac{|\mathbf e_i\mathbf V_{1:c}^T\mathbf U\mathbf U^T\mathbf V_{1:c}\mathbf e_i|^2}{\|\mathbf U^T\mathbf V_{1:c}\mathbf V_{1:c}^T\mathbf e_i\|^2}
    = \frac{|\mathbf e_i^T\mathbf V_{1:c}\mathbf V_{1:c}^T\mathbf e_i|^2}{\mathbf e_i^T\mathbf V_{1:c}\mathbf V_{1:c}^T\mathbf e_i}= \|\mathbf V_{1:c}^T\mathbf e_i\|^2,
\end{aligned}
\end{equation}
which gives
\begin{equation}
    \theta_1=\arccos(\|\mathbf V_{1:c}^T\mathbf e_i\|).
\end{equation}
For $\theta_2$, as rescaling does not change the direction of a vector, we can rewrite $\theta_2$ as
\begin{equation}\label{eq:theta_2_rewrite}
    \theta_2=\measuredangle(\mathbf U^T\zeta,\Lambda^{-1}(\widehat{\boldsymbol\Sigma}_0)\mathbf U^T\zeta),
\end{equation}
where
\begin{equation}
     \zeta= \frac{\mathbf V_{1:c}\mathbf V_{1:c}^T\mathbf e_i}{\|\mathbf V_{1:c}\mathbf V_{1:c}^T\mathbf e_i\|}.
\end{equation}
Note that $\zeta$ is a unit-length random vector and is independent of $\mathbf U$ due to the independency between $\mathbf V_{1:c}$ and $\mathbf U$. Then, we have
\begin{equation}\label{eq:theta2}
\begin{aligned}
    \cos^2(\theta_2)
    = \frac{|\zeta^T\mathbf U\Lambda^{-1}(\widehat{\boldsymbol\Sigma}_0)\mathbf U^T\zeta|^2}{\|\Lambda^{-1}(\widehat{\boldsymbol\Sigma}_0)\mathbf U^T\zeta\|^2}=\frac{(\zeta^T\mathbf U\Lambda^{-1}(\widehat{\boldsymbol\Sigma}_0)\mathbf U^T\zeta)^2}{\zeta^T\mathbf U\Lambda^{-2}(\widehat{\boldsymbol\Sigma}_0)\mathbf U^T\zeta}.
\end{aligned}
\end{equation}
We have known, from Lemma \ref{lm:property_Sigma0}, $\mathbf U$ is uniformly distributed on the set of all orthonormal matrices in $\mathbb R^{D\times D}$, and $\zeta$ is a unit-length random vector independent of $\mathbf U$. Thus, $\xi=\mathbf U^T\zeta$ must be a unit-length random vector uniformly distributed on the unit sphere $\mathbb S^{D-1}$. Finally, (\ref{eq:theta2}) gives
\begin{equation}
    \theta_2=\arccos\left(\xi^T\Lambda^{-1}(\widehat{\boldsymbol\Sigma}_0)\xi\Big/
    \sqrt{\xi^T\Lambda^{-2}(\widehat{\boldsymbol\Sigma}_0)\xi}\right).
\end{equation}
This completes the proof.

\subsection{Proof of Lemma \ref{lm:property_Sigma0}}
Since $\widehat{\boldsymbol\Sigma}_0=\mathbf X^T\widehat{\boldsymbol\Sigma}\mathbf X$ is a normalized sample covariance, wherein $\mathbf X^T\boldsymbol\Sigma\mathbf X=\mathbf I$, we have
\begin{align}
\widehat{\boldsymbol\Sigma}_0 &= \frac1N\sum_{i=1}^{c+1}\sum_{j=1}^n(\mathbf x_j^i-\bar{\mathbf x}_i)(\mathbf x_j^i-\bar{\mathbf x}_i)^T,
\end{align}
where $\mathbf x_j^i$ is sampled from some $\mathcal N(\boldsymbol\mu_i,\mathbf I)$ and $\bar{\mathbf x}_i$ is the sample mean. Letting $\mathbf z_j^i=\mathbf x_j^i-\boldsymbol\mu_i$, which implies $\mathbf z_j^i$ is sampled from the standard Gaussian distribution $\mathcal N(0,\mathbf I)$, and $\bar{\mathbf z}^i=\bar{\mathbf x}^i-\boldsymbol\mu_i$, then $\widehat{\boldsymbol\Sigma}_0$ can be rewritten as
\begin{align}\label{eq:sample_covariance}
\widehat{\boldsymbol\Sigma}_0 &= \frac1N\sum_{i=1}^{c+1}\sum_{j=1}^n(\mathbf z_j^i-\bar{\mathbf z}^i)(\mathbf z_j^i-\bar{\mathbf z}^i)^T,
\end{align}
One property of $\widehat{\boldsymbol\Sigma}_0$ in (\ref{eq:sample_covariance}) is that, as a random variable, its distribution is invariant to orthogonal similarity transformation, i.e., $\widehat{\boldsymbol\Sigma}_0$ and $\mathbf O\widehat{\boldsymbol\Sigma}_0\mathbf O^T$, wherein $\mathbf U^T\mathbf U=\mathbf I$, have the same distribution. This is due to the fact that $\mathbf O^T\widehat{\boldsymbol\Sigma}_0\mathbf O$ corresponds to (\ref{eq:sample_covariance}) in the case of replacing $\mathbf z_j^i$ by $\mathbf O\mathbf z_j^i$ while $\mathbf O\mathbf z_j^i$ has the same distribution with $\mathbf z_j^i$, i.e., the standard Gaussian distribution $\mathcal N(0,\mathbf I)$. Then, according to Theorem 3.2 in \cite{edelman1989eigenvalues}, the invariant property to orthogonal similarity transformation implies that the distribution of $\widehat{\boldsymbol\Sigma}_0$ is independent of its eigenvectors $\mathbf U$ but only depends on its eigenvalues $\Lambda(\widehat{\boldsymbol\Sigma}_0)$, and $\mathbf U$ is a random matrix uniformly distributed on the set of all possible orthonormal matrices in $\mathbb R^{D\times D}$. This completes the statements 1) and 2) in Lemma \ref{lm:property_Sigma0}.

Further, (\ref{eq:sample_covariance}) can be rewritten as
\begin{equation}\label{eq:Sigma0_pertur}
\begin{aligned}
    \widehat{\boldsymbol\Sigma}_0 &= \frac1N\sum_{i=1}^{c+1}\sum_{j=1}^{n}\mathbf z_j^i\mathbf z_j^{iT}- \frac1{c+1}\sum_{i=1}^{c+1}\bar{\mathbf z}^i\bar{\mathbf z}^{iT}
    =\frac1N\sum_{i=1}^{c+1}\sum_{j=1}^{n}\mathbf z_j^i\mathbf z_j^{iT}- \frac1{(c+1)n}\sum_{i=1}^{c+1}\sqrt{n}\bar{\mathbf z}^i\sqrt{n}\bar{\mathbf z}^{iT}\\
    &=\frac1N\mathbf G_1\mathbf G_1^T - \frac1N\mathbf G_2\mathbf G_2^T
    = T_1 + T_2.
    \end{aligned}
\end{equation}
where $\mathbf G_1\in\mathbb R^{D\times N}$ and $\mathbf G_2\in\mathbb R^{D\times (c+1)}$. For the first term $T_1=\frac1N\mathbf G_1\mathbf G_1^T$, by Proposition \ref{propos:MP_Law1}, we know that the empirical distribution of its eigenvalues converges almost surely to $F_{\gamma}(\lambda)$ with density,
\begin{equation}\label{eq:tmp00}
    dF_{\gamma}(\lambda)= \frac{\sqrt{(\lambda_+ - \lambda)(\lambda-\lambda_-)}}{2\pi\gamma\lambda}d\lambda,
\end{equation}
where $\gamma = \lim D/N$ and
\begin{equation}
    \lambda_+ = (1+\sqrt{\gamma})^2 \mbox{~and~} \lambda_- = (1-\sqrt{\gamma})^2.
\end{equation}
For the second term $T_2=\frac1N\mathbf G_2\mathbf G_2^T$, clearly it has finite rank $c+1$. According to \cite{TaoRMT}, a finite rank perturbation does not effect the convergence of the empirical spectral distribution, i.e., $\lim F_N(\lambda(T_1+T2))=\lim F_N(\lambda(T_1))=F_{\gamma}(\lambda)$.
This completes the proof.

\subsection{Proof of Lemma \ref{lm:minus1moment}}
The condition that $\xi$ is a unit-length random vector uniformly distributed on the unit sphere $\mathbb S^{D-1}$ can be replaced by $\xi\in\mathbb R^D$ with entries independently sampled from $\mathcal N(0,1/D)$. This is because, in the later case, $\xi/\|\xi\|$ is uniformly distributed on $\mathbb S^{D-1}$, and $\|\xi\|^2\overset{a.s.}\longrightarrow1$ due to the Strong Law of Large Numbers.

For (\ref{eq:minus1moment}), we divide the proof into two steps. First, we show that $\xi^T\Lambda^{-1}(\widehat{\boldsymbol\Sigma}_0)\xi\overset{a.s.}\longrightarrow\int \lambda^{-1} dF_{\gamma}(\lambda)$, and then we calculate the integral.

\noindent i)
Recall $\lambda_- =(1-\sqrt{\gamma})^2$, and let $\overline{\Lambda}^{-1}(\widehat{\boldsymbol\Sigma}_0)=\mathrm{diag}(\min\{\lambda_-,\lambda_i^{-1}(\widehat{\boldsymbol\Sigma}_0)\})$, i.e., a truncated version of $\Lambda^{-1}(\widehat{\boldsymbol\Sigma}_0)$ by clamping $\lambda_i^{-1}(\widehat{\boldsymbol\Sigma}_0)$ to be $\lambda_-^{-1}$ if $\lambda_i^{-1}(\widehat{\boldsymbol\Sigma}_0)\ge\lambda_-^{-1}$. Then, we divide the left-hand side of (\ref{eq:minus1moment}) into three terms
\begin{equation}\label{eq:Term1}
    \xi^T\Lambda^{-1}(\widehat{\boldsymbol\Sigma}_0)\xi -
    \xi^T\overline\Lambda^{-1}(\widehat{\boldsymbol\Sigma}_0)\xi,
\end{equation}

\begin{equation}\label{eq:Term2}
    \xi^T\overline\Lambda^{-1}(\widehat{\boldsymbol\Sigma}_0)\xi - \frac1D\mathrm{Tr}(\overline\Lambda^{-1}(\widehat{\boldsymbol\Sigma}_0)),
\end{equation}
and
\begin{equation}\label{eq:Term3}
    \frac1D\mathrm{Tr}(\overline\Lambda^{-1}(\widehat{\boldsymbol\Sigma}_0)) - \int\lambda^{-1}dF_{\gamma}(\lambda).
\end{equation}
We show that all the three terms converge almost surely to zero.

For the first term (\ref{eq:Term1}), we have
\begin{equation}\label{eq:T11}
\begin{aligned}
0\le&\xi^T(\Lambda^{-1}(\widehat{\boldsymbol\Sigma}_0)-\overline\Lambda^{-1}(\widehat{\boldsymbol\Sigma}_0))\xi\\
    \le&\|\xi\|^2\max\{0,\lambda_{\min}^{-1}(\widehat{\boldsymbol\Sigma}_0)-\lambda_-^{-1}\}.\\
\end{aligned}
\end{equation}
By the same argument in the proof of Lemma \ref{lm:property_Sigma0}, we know that
\begin{equation}
    \lim\lambda_{min}(\widehat{\boldsymbol\Sigma}_0)=\lim\lambda_{min}\left(\frac1N\sum_{i=1}^{c+1}\sum_{j=1}^{n}\mathbf z_j^i\mathbf z_j^{iT}\right) = \left(\lim\frac1{\sqrt{N}}\sigma_{min}(\mathbf Z)\right)^2,
\end{equation}
where $\mathbf Z=[\mathbf z_1^1,...,\mathbf z_n^{c+1}]\in\mathbb R^{D\times N}$, with entries independently sampled from $\mathcal N(0,1)$. By Proposition \ref{lm:largest_singularvalue_GRM}, we have $\lim\frac1{\sqrt{N}}\sigma_{min}(\mathbf Z) = 1-\sqrt{\gamma}$, and thus $\lambda_{min}(\widehat{\boldsymbol\Sigma}_0) \overset{a.s.}\longrightarrow (1-\sqrt{\gamma})^2= {\lambda}_-$. Accordingly,
\begin{equation}\label{eq:T12}
    \max\{0,\lambda_{\min}^{-1}(\widehat{\boldsymbol\Sigma}_0)-\lambda_-^{-1}\}\overset{a.s.}\longrightarrow0.
\end{equation}
Then, by $\|\xi\|^2\overset{a.s.}\longrightarrow1$, (\ref{eq:T11}) and (\ref{eq:T12}), we have
\begin{equation}
    \xi^T\Lambda^{-1}(\widehat{\boldsymbol\Sigma}_0)\xi-\xi^T\overline\Lambda^{-1}(\widehat{\boldsymbol\Sigma}_0)\xi\overset{a.s.}\longrightarrow0.
\end{equation}

For the second term (\ref{eq:Term2}), since $\|\overline\Lambda^{-1}(\widehat{\boldsymbol\Sigma}_0)\|\le\lambda_-$ for all $D$, i.e., it is uniformly bounded, we apply Theorem 3.4 in \cite{tulino2004random} and get
\begin{equation}
    \xi^T\overline\Lambda_{\alpha}^{-1}(\widehat{\boldsymbol\Sigma}_0)\xi - \frac1D\mathrm{Tr}(\overline\Lambda_{\alpha}^{-1}(\widehat{\boldsymbol\Sigma}_0))
    \overset{a.s.}\longrightarrow 0.
\end{equation}

For the third term (\ref{eq:Term3}), since $dF_{\gamma}(\lambda)$ is nonzero only on $[\lambda_-,\lambda_+]$, it is sufficient to examine
\begin{equation}
\begin{aligned}
    &\frac1D\mathrm{Tr}(\overline\Lambda^{-1}(\widehat{\boldsymbol\Sigma}_0)) - \int\lambda^{-1}dF_{\gamma}(\lambda)
    =\int_0^{\infty}\min(\lambda_-,\lambda^{-1})dF_N(\lambda) - \int_{\lambda_-}^{\lambda_+}\lambda^{-1}dF_{\gamma}(\lambda)\\
    =&\int_{\lambda_-}^{\lambda_+}\lambda^{-1}d(F_N(\lambda) -F_{\gamma}(\lambda)) + \lambda_-^{-1}\int_0^{\lambda_-} dF_N(\lambda) + \int_{\lambda_+}^{\infty} \lambda^{-1}dF_N(\lambda).\\
\end{aligned}
\end{equation}
Sine $F_N(\lambda)\overset{a.s.}\longrightarrow F_{\gamma}(\lambda)$ and $\lambda^{-1}$ is bounded on $[{\lambda_-},{\lambda_+}]$, it holds \cite{ConvergenceOfMeasure:Billingsley}
\begin{equation}
    \int_{\lambda_-}^{\lambda_+}\lambda^{-1}d(F_N(\lambda) -F_{\gamma}(\lambda))\overset{a.s.}\longrightarrow 0.
\end{equation}
Further, sine $F_{\gamma}(\lambda_-)=0$ and $F_{\gamma}(\lambda_+)=1$, it holds
\begin{equation}
    \int_0^{\lambda_-} dF_N(\lambda)= F_N(\lambda_-)\overset{a.s.}\longrightarrow F_{\gamma}(\lambda_-)=0,
\end{equation}
and
\begin{equation}
    0\le\int_{\lambda_+}^{\infty}\lambda^{-1} dF_N(\lambda)\le\lambda_+^{-1} (1-F_N(\lambda_+))\overset{a.s.}\longrightarrow \lambda_+^{-1} (1-F_{\gamma}(\lambda_+))=0.
\end{equation}
Thus,
\begin{equation}
   \frac1D\mathrm{Tr}(\overline\Lambda_{\alpha}^{-1}(\widehat{\boldsymbol\Sigma}_0)) - \int\lambda^{-1}dF_{\gamma}(\lambda)\overset{a.s.}\longrightarrow0.
\end{equation}

\noindent ii) We now calculate the integral
\begin{equation}\label{eq:integral}
    I=\int \lambda^{-1} dF_{\gamma}(\lambda)=\int_{\lambda_-}^{\lambda_+}\frac{\sqrt{(\lambda_+-\lambda)(\lambda - \lambda_-)}}{2\pi\gamma\lambda^2}d\lambda
\end{equation}
where $\lambda_+ = (1+\sqrt{\gamma})^2$ and $\lambda_-= (1-\sqrt{\gamma})^2$.

Letting $\lambda = 1+\gamma-2\sqrt{\gamma}\cos x$, $x\in[0,\pi]$ and substituting it into (\ref{eq:integral}), we have
\begin{equation}
    I= \frac2\pi\int_{0}^{\pi}\frac{\sin^2 x}{(1+\gamma-2\sqrt{\gamma}\cos x)^2}dx.
\end{equation}
Further, letting $t=\tan\frac{x}2$, we have
\begin{equation}\label{eq:int_0}
\begin{aligned}
    I&= \frac2\pi\int_{0}^{\infty}\frac{\left(\frac{2t}{1+t^2}\right)^2}{\left(1+\gamma-2\sqrt{\gamma}\frac{1-t^2}{1+t^2}\right)^2}\frac{2}{1+t^2}dt
    =\frac{16}\pi\int_{0}^{\infty}\frac{t^2}{\left((1+\gamma)(t^2+1)-2\sqrt{\gamma}(1-t^2)\right)^2}\frac{1}{1+t^2}dt\\
    &=\frac{16}{\pi}\int_{0}^{\infty}\frac{t^2}{\left((1+\sqrt{\gamma})^2t^2 + (1-\sqrt{\gamma})^2\right)^2}\frac{1}{1+t^2}dt
    =\frac{16}{\pi(1+\sqrt{\gamma})^4}\int_{0}^{\infty}\frac{t^2}{\left(t^2 + \left(\frac{1-\sqrt{\gamma}}{1+\sqrt{\gamma}}\right)^2\right)^2}\frac{1}{1+t^2}dt.
\end{aligned}
\end{equation}
Letting $\alpha = \frac{1-\sqrt{\gamma}}{1+\sqrt{\gamma}}$ and by partial fraction, we have
\begin{equation}\label{eq:int_part}
\begin{aligned}
    \int_{0}^{\infty}\frac{t^2}{\left(t^2 + \alpha^2\right)^2}\frac{1}{1+t^2}dt
    &=\int_{0}^{\infty}\frac{-\frac1{(1-\alpha^2)^2}}{t^2+1}dt + \int_{0}^{\infty}\frac{\frac1{(1-\alpha^2)^2}}{t^2+\alpha^2}dt + \int_{0}^{\infty}\frac{-\frac{\alpha^2}{(1-\alpha^2)}}{(t^2+\alpha^2)^2}dt.
\end{aligned}
\end{equation}
Denoting by $I_1$, $I_2$ and $I_3$ the terms in the righthand side of (\ref{eq:int_part}), we have
\begin{equation}\label{eq:int_part1}
\begin{aligned}
    I_1 &=\int_{0}^{\infty}\frac{-\frac1{(1-\alpha^2)^2}}{t^2+1}dt
    = \frac{-1}{(1-\alpha^2)^2}\int_0^{\infty}d\arctan t = \frac{-\pi}{2(1-\alpha^2)^2},
\end{aligned}
\end{equation}
\begin{equation}\label{eq:int_part2}
\begin{aligned}
    I_2 &= \int_{0}^{\infty}\frac{\frac1{(1-\alpha^2)^2}}{t^2+\alpha^2}dt
    =\frac{1}{\alpha(1-\alpha^2)^2}\int_0^{\infty}d\arctan \frac{t}\alpha = \frac{\pi}{2\alpha(1-\alpha^2)^2},
\end{aligned}
\end{equation}
\begin{equation}\label{eq:int_part3}
\begin{aligned}
    I_3 &= \int_{0}^{\infty}\frac{-\frac{\alpha^2}{(1-\alpha^2)}}{(t^2+\alpha^2)^2}dt
    = \frac{-1}{2(1-\alpha^2)}\int_{0}^{\infty}d\frac{t}{t^2+\alpha^2} + \frac{-1}{2(1-\alpha^2)}\int_{0}^{\infty}\frac{1}{t^2+\alpha^2}dt\\
    & = 0+ \frac{-\pi}{4\alpha(1-\alpha^2)}=\frac{-\pi}{4\alpha(1-\alpha^2)}.
\end{aligned}
\end{equation}

Combining (\ref{eq:int_0}) to (\ref{eq:int_part3}) and noticing $\alpha = \frac{1-\sqrt{\gamma}}{1+\sqrt{\gamma}}$, we get
\begin{equation}
\begin{aligned}
    I &= \frac{16}{\pi(1+\sqrt{\gamma})^4} \left(\frac{-\pi}{2(1-\alpha^2)^2} + \frac{\pi}{2\alpha(1-\alpha^2)^2} +\frac{-\pi}{4\alpha(1-\alpha^2)}\right)\\
    &=\frac{16}{\pi(1+\sqrt{\gamma})^4}\frac{\pi}{4\alpha(1+\alpha)^2}
    =\frac{1}{1-\gamma}.
\end{aligned}
\end{equation}
This completes the proof of (\ref{eq:minus1moment}).

For (\ref{eq:minus2moment}), by the same strategy as used in the proof of (\ref{eq:minus1moment}), we have $\xi^T\Lambda^{-2}(\widehat{\boldsymbol\Sigma}_0)\xi\overset{a.s.}\longrightarrow\int \lambda^{-2} dF_{\gamma}(\lambda)$. Below, we calculate the integral.
\begin{equation}\label{eq:integral_2}
    I=\int \lambda^{-2} dF_{\gamma}(\lambda)=\int_{\lambda_-}^{\lambda_+}\frac{\sqrt{(\lambda_+-\lambda)(\lambda - \lambda_-)}}{2\pi\gamma\lambda^3}d\lambda,
\end{equation}
where $\lambda_+ = (1+\sqrt{\gamma})^2$ and $\lambda_-= (1-\sqrt{\gamma})^2$.
Letting $\lambda = 1+\gamma-2\sqrt{\gamma}\cos x$, $x\in[0,\pi]$ and substituting it into (\ref{eq:integral}), we have
\begin{equation}
    I= \frac2\pi\int_{0}^{\pi}\frac{\sin^2 x}{(1+\gamma-2\sqrt{\gamma}\cos x)^3}dx.
\end{equation}
Further, letting $t=\tan\frac{x}2$, we have
\begin{equation}\label{eq:int_0_2}
\begin{aligned}
    I&= \frac2\pi\int_{0}^{\infty}\frac{\left(\frac{2t}{1+t^2}\right)^2}{\left(1+\gamma-2\sqrt{\gamma}\frac{1-t^2}{1+t^2}\right)^3}\frac{2}{1+t^2}dt
    =\frac{16}\pi\int_{0}^{\infty}\frac{t^2}{\left((1+\gamma)(t^2+1)-2\sqrt{\gamma}(1-t^2)\right)^3}dt\\
    &=\frac{16}{\pi}\int_{0}^{\infty}\frac{t^2}{\left((1+\sqrt{\gamma})^2t^2 + (1-\sqrt{\gamma})^2\right)^3}dt
    =\frac{16}{\pi(1+\sqrt{\gamma})^6}\int_{0}^{\infty}\frac{t^2}{\left(t^2 + \left(\frac{1-\sqrt{\gamma}}{1+\sqrt{\gamma}}\right)^2\right)^3}dt.
\end{aligned}
\end{equation}
Letting $\alpha = \frac{1-\sqrt{\gamma}}{1+\sqrt{\gamma}}$, we have
\begin{equation}\label{eq:int_part_2}
\begin{aligned}
    \int_{0}^{\infty}\frac{t^2}{\left(t^2 + \alpha^2\right)^3}dt
    =&-\frac14\int_{0}^{\infty}d\frac{t}{(t^2+\alpha^2)^2} +\frac14\int_{0}^{\infty}\frac{1}{(t^2+\alpha^2)^2}dt
    =\frac\pi{16\alpha^3}.
\end{aligned}
\end{equation}
Thus, by $\alpha = \frac{1-\sqrt{\gamma}}{1+\sqrt{\gamma}}$, we get $I =\frac{16}{\pi(1+\sqrt{\gamma})^6} \frac\pi{16\alpha^3}=\frac1{(1-\gamma)^3}$.
This completes the proof of (\ref{eq:minus2moment}).

\subsection{Proof of Lemma \ref{lm:eigenspace_S0}}
By Lemmas \ref{lm:double_diagonal} and \ref{lm:gen_dis_pow}, $\widehat{\mathbf S}_0$ is an estimate of $\mathbf X^T\mathbf S\mathbf X=\boldsymbol\Lambda_0=\mbox{diag}(\boldsymbol\lambda_1,...,\boldsymbol\lambda_c,0,...,0)$. Suppose the original distributions of the $c+1$ classes are $\mathcal N(\boldsymbol\mu_i,\boldsymbol\Sigma)$ and the between-class scatter matrix is $\mathbf S$. Then, $\boldsymbol\Lambda_0$ should be the between-class scatter matrix of an equivalent problem with distributions $\mathcal N(\boldsymbol\mu_i',\mathbf I)$, wherein $\boldsymbol\mu_i'=\mathbf X^{T}\boldsymbol\mu_i$. Therefore, $\boldsymbol\Lambda_0=\frac{1}{c+1}\sum_{i=1}^{c+1}(\boldsymbol\mu_i'-\boldsymbol\mu')(\boldsymbol\mu_i'-\boldsymbol\mu')^T$, with $\boldsymbol\mu'=\frac{1}{c+1}\sum_{i=1}^{c+1}\boldsymbol\mu_i'$. Letting $\mathbf M=[\boldsymbol\mu_1',...,\boldsymbol\mu_{c+1}']$ and $\mathbf E\in\mathbb R^{(c+1)\times (c+1)}$ with all entries equal to $\frac1{c+1}$, we have $\boldsymbol\Lambda_0=\frac1{c+1}\mathbf M(\mathbf I-\mathbf E)(\mathbf I-\mathbf E)^T\mathbf M^T$. Similarly, we have $\widehat{\mathbf S}_0=\frac1{c+1}\widehat{\mathbf M}(\mathbf I-\mathbf E)(\mathbf I-\mathbf E)^T\widehat{\mathbf M}^T$, where $\widehat{\mathbf M}=[\widehat{\boldsymbol\mu}_1',...,\widehat{\boldsymbol\mu}_{c+1}']$ and $\widehat{\boldsymbol\mu}_1'$ is an estimate of ${\boldsymbol\mu}_1'$. As there are $n$ training examples per class, we have $\widehat{\mathbf M} = \mathbf M + \mathbf X$, where the entries of $\mathbf X\in\mathbb R^{D\times (c+1)}$ are i.i.d. samples from $\mathcal N(0,1/n)$.

Note that the nonzero diagonal entries of $\boldsymbol\Lambda_0$ are $\boldsymbol\lambda_i$, $i=1,2,...,c$, which are actually eigenvalues of $\boldsymbol\Lambda_0$, associated with eigenvectors $\mathbf e_i$, $i=1,2,...,c$. Thus, $\boldsymbol\Lambda_0=\frac1{c+1}\mathbf M(\mathbf I-\mathbf E)(\mathbf I-\mathbf E)^T\mathbf M^T$ implies that ${\mathbf M}(\mathbf I-\mathbf E)$ has singular values $\sqrt{(c+1)\boldsymbol\lambda_i}$, $i=1,2,...,c$ and left singular vectors $\mathbf I_{1:c}=[\mathbf e_1,...,\mathbf e_c]$. Denoting by $\mathbf Q\in\mathbb R^{(c+1)\times c}$ the right singular vectors of ${\mathbf M}(\mathbf I-\mathbf E)$, $\mathbf Q^T\mathbf Q=\mathbf I_c$, we have
\begin{equation}
    \mathbf M(\mathbf I-\mathbf E)\mathbf Q = \left[\sqrt{(c+1)\boldsymbol\lambda_1}\mathbf e_1,...,\sqrt{(c+1)\boldsymbol\lambda_c}\mathbf e_c\right].
\end{equation}
Consequently, by $\widehat{\mathbf M}=\mathbf M + \mathbf X$, we have
\begin{equation}
\begin{aligned}
     \widehat{\mathbf M}(\mathbf I-\mathbf E)\mathbf Q &= \left[\sqrt{(c+1)\boldsymbol\lambda_1}\mathbf e_1,...,\sqrt{(c+1)\boldsymbol\lambda_c}\mathbf e_c\right] + \mathbf X(\mathbf I - \mathbf E)\mathbf Q
      = [\boldsymbol\xi_1,...,\boldsymbol\xi_c],
\end{aligned}
\end{equation}
where
\begin{align}
\boldsymbol\xi_i = \sqrt{(c+1)\boldsymbol\lambda_i}\mathbf e_i + \mathbf X(\mathbf I - \mathbf E)\mathbf Q_i,~i=1,2,...,c.
\end{align}
Then, by $\widehat{\mathbf S}_0=\frac1{c+1}\widehat{\mathbf M}(\mathbf I-\mathbf E)(\mathbf I-\mathbf E)^T\widehat{\mathbf M}^T$, we have for the first $c$ eigenvectors $\mathbf V_{1:c}$ of $\widehat{\mathbf S}_0$ that

\begin{equation}
\begin{aligned}
    \mathbf V_{1:c} = \mathcal R(\widehat{\mathbf M}(\mathbf I-\mathbf E))= \mathcal R(\widehat{\mathbf M}(\mathbf I-\mathbf E)\mathbf Q)
     = \mathcal R([\boldsymbol\xi_1,...,\boldsymbol\xi_c]).
\end{aligned}
\end{equation}
Accordingly,
\begin{equation}\label{eq:Ve}
\begin{aligned}
    \|\mathbf V_{1:c}^T\mathbf e_i\|^2 &= \|\mathcal R^T([\boldsymbol\xi_1,...,\boldsymbol\xi_c])\mathbf e_i\|^2
    \ge\|\mathcal R^T(\boldsymbol\xi_i)\mathbf e_i\|^2
    =\frac1{\|\boldsymbol\xi_i\|^2}|\boldsymbol\xi_i^T\mathbf e_i|^2\\
    &=\frac{|\mathbf e_i^T\sqrt{(c+1)\boldsymbol\lambda_i}\mathbf e_i + \mathbf e_i^T\mathbf X(\mathbf I - \mathbf E)\mathbf Q_i|^2}{\|\sqrt{(c+1)\boldsymbol\lambda_i}\mathbf e_i + \mathbf X(\mathbf I - \mathbf E)\mathbf Q_i\|^2}\\
    &\ge\frac{{(c+1)\boldsymbol\lambda_i} + |\mathbf e_i^T\mathbf X(\mathbf I - \mathbf E)\mathbf Q_i|^2-2\sqrt{(c+1)\boldsymbol\lambda_i}|\mathbf e_i^T\mathbf X(\mathbf I - \mathbf E)\mathbf Q_i|}{{(c+1)\boldsymbol\lambda_i} + \|\mathbf X(\mathbf I - \mathbf E)\mathbf Q_i\|^2 + 2\sqrt{(c+1)\boldsymbol\lambda_i}\mathbf e_i^T\mathbf X(\mathbf I - \mathbf E)\mathbf Q_i}.
\end{aligned}
\end{equation}
It can be verified that as $N=(c+1)n\longrightarrow\infty$
\begin{equation}\label{eq:eXQ}
    |\mathbf e_i^T\mathbf X(\mathbf I - \mathbf E)\mathbf Q_i|\le\|\mathbf e_i^T\mathbf X\|=\sqrt{\sum_{j=1}^{c+1}\mathbf X_{ij}^2}\overset{a.s.}\longrightarrow0,
\end{equation}
where the inequality is due to $\|(\mathbf I - \mathbf E)\mathbf Q_i\|\le\|(\mathbf I - \mathbf E)\|\|\mathbf Q_i\|\le1$ and the limit is because $\mathbf X_{ij}$ follows the distribution $\mathcal N(0,\frac1n)$.

In addition, by Proposition \ref{lm:largest_singularvalue_GRM} and letting $\mathbf G=\sqrt{n}\mathbf X$, we have
\begin{equation}\label{eq:s_gamma}
    \|\mathbf X\|=\frac{1}{\sqrt{n}}\|\mathbf G\|\overset{a.s.}\longrightarrow\sqrt{\frac{D}n}=\sqrt{\frac{(c+1)D}N}\longrightarrow\sqrt{(c+1)\gamma}.
\end{equation}
Thus,
\begin{equation}\label{eq:XIEQ}
    \|\mathbf X(\mathbf I - \mathbf E)\mathbf Q_i\|\le\|\mathbf X\|\overset{a.s.}\longrightarrow\sqrt{(c+1)\gamma}.
\end{equation}

Combining (\ref{eq:Ve}), (\ref{eq:eXQ}) and (\ref{eq:XIEQ}), we obtain
\begin{equation}
    \lim_{D/N\longrightarrow\gamma}\|\mathbf V_{1:c}^T\mathbf e_i\|^2\ge\frac{\boldsymbol\lambda_i}{\boldsymbol\lambda_i + \gamma}, ~a.s.
\end{equation}
This completes the proof.

\subsection{Proof of Corollary \ref{coly:generalization_bound}}
Recall that
\begin{align}
  P =  0.5\Phi \left\{-\frac{\widehat{\mathbf w}_1^T\boldsymbol\mu_1-0.5\widehat{\mathbf w}_1^T(\widehat{\boldsymbol\mu}_1+\widehat{\boldsymbol\mu}_2)}{\sqrt{\widehat{\mathbf w}_1^T\boldsymbol\Sigma\widehat{\mathbf w}_1}}\right\}+0.5\Phi \left\{-\frac{0.5\widehat{\mathbf w}_1^T(\widehat{\boldsymbol\mu}_1+\widehat{\boldsymbol\mu}_2)-\widehat{\mathbf w}_1^T\boldsymbol\mu_2}{\sqrt{\widehat{\mathbf w}_1^T\boldsymbol\Sigma\widehat{\mathbf w}_1}}\right\},\label{eq:flda_gener_error}
\end{align}
assumed $\widehat{\mathbf w}_1^T(\boldsymbol\mu_1-\boldsymbol\mu_2)\ge0$. First, we have
\begin{equation}
\begin{aligned}
  -\frac{\widehat{\mathbf w}_1^T\boldsymbol\mu_1-0.5\widehat{\mathbf w}_1^T(\widehat{\boldsymbol\mu}_1+\widehat{\boldsymbol\mu}_2)}{\sqrt{\widehat{\mathbf w}_1^T\boldsymbol\Sigma\widehat{\mathbf w}_1}}
   = &-0.5\frac{\widehat{\mathbf w}_1^T(\boldsymbol\mu_1-\boldsymbol\mu_2)}{\sqrt{\widehat{\mathbf w}_1^T\boldsymbol\Sigma\widehat{\mathbf w}_1}} +
   0.5\frac{\widehat{\mathbf w}_1^T((\widehat{\boldsymbol\mu}_1+\widehat{\boldsymbol\mu}_2)-(\boldsymbol\mu_1+\boldsymbol\mu_2))}{\sqrt{\widehat{\mathbf w}_1^T\boldsymbol\Sigma\widehat{\mathbf w}_1}}\\
   =&-\sqrt{\frac{\widehat{\mathbf w}_1^T\mathbf S\widehat{\mathbf w}_1}{{\widehat{\mathbf w}_1^T\boldsymbol\Sigma\widehat{\mathbf w}_1}}}+0.5\frac{\widehat{\mathbf w}_1^T((\widehat{\boldsymbol\mu}_1+\widehat{\boldsymbol\mu}_2)-(\boldsymbol\mu_1+\boldsymbol\mu_2))}{\sqrt{\widehat{\mathbf w}_1^T\boldsymbol\Sigma\widehat{\mathbf w}_1}}\\
   =&-\sqrt{\boldsymbol\delta_1\boldsymbol\lambda_1} + 0.5T,
\end{aligned}
\end{equation}
and similarly
\begin{equation}
\begin{aligned}
  -\frac{0.5\widehat{\mathbf w}_1^T(\widehat{\boldsymbol\mu}_1+\widehat{\boldsymbol\mu}_2)-\widehat{\mathbf w}_1^T\boldsymbol\mu_2}{\sqrt{\widehat{\mathbf w}_1^T\boldsymbol\Sigma\widehat{\mathbf w}_1}}
   = &-0.5\frac{\widehat{\mathbf w}_1^T(\boldsymbol\mu_1-\boldsymbol\mu_2)}{\sqrt{\widehat{\mathbf w}_1^T\boldsymbol\Sigma\widehat{\mathbf w}_1}} -
   0.5\frac{\widehat{\mathbf w}_1^T((\widehat{\boldsymbol\mu}_1+\widehat{\boldsymbol\mu}_2)-(\boldsymbol\mu_1+\boldsymbol\mu_2))}{\sqrt{\widehat{\mathbf w}_1^T\boldsymbol\Sigma\widehat{\mathbf w}_1}}\\
   =&-\sqrt{\boldsymbol\delta_1\boldsymbol\lambda_1} - 0.5T,
\end{aligned}
\end{equation}

As long as $T\overset{a.s.}\longrightarrow0$, we have by Theorem \ref{thm:generalization_bound} that
\begin{equation}
  P=\Phi(-\sqrt{\boldsymbol\delta_1\boldsymbol\lambda_1})\le\Phi(-\varrho\sqrt{\boldsymbol\lambda_1})
\end{equation}
with
\begin{equation}
  \varrho={\max}\big\{\cos(\arccos(\sqrt{{\boldsymbol\lambda_i}/(\boldsymbol\lambda_i + \gamma)}) + \arccos(\sqrt{1-\gamma})),0\big\}.
\end{equation}

Below, we verify that it indeed holds
\begin{align}\label{eq:T}
  T=\frac{\widehat{\mathbf w}_1^T((\widehat{\boldsymbol\mu}_1-\boldsymbol\mu_1)+(\widehat{\boldsymbol\mu}_2 -\boldsymbol\mu_2))}{\sqrt{\widehat{\mathbf w}_1^T\boldsymbol\Sigma\widehat{\mathbf w}_1}}\overset{a.s.}\longrightarrow0.
\end{align}
By using similar strategy in the proof of Lemma \ref{lm:gen_dis_pow}, in particular (\ref{eq:Q_useful}), we have $\widehat{\mathbf w}_1=\mathbf X\mathbf q$, wherein $\mathbf X$ satisfies $\mathbf X^T\boldsymbol\Sigma\mathbf X=\mathbf I$ and
\begin{align}\label{eq:qq}
  \mathbf q = a\mathbf U^T\Lambda^{-1}(\widehat{\boldsymbol\Sigma}_0)\mathbf U\mathbf X^T(\widehat{\boldsymbol\mu}_1-\widehat{\boldsymbol\mu}_2),~\mbox{for some}~a\ne0,  ~
\end{align}
since $\mathbf X(\widehat{\boldsymbol\mu}_1-\widehat{\boldsymbol\mu}_2)$ is the first eigenvector of the normalized sample between-scatter matrix $\widehat{\mathbf S}_0=\mathbf X^T\widehat{\mathbf S}\mathbf X$. Substituting (\ref{eq:qq}) into $T$, we have

\begin{equation}
\begin{aligned}
  T&=\frac{(\widehat{\boldsymbol\mu}_1-\widehat{\boldsymbol\mu}_2)^T\mathbf X\mathbf U^T\Lambda^{-1}(\widehat{\boldsymbol\Sigma}_0)\mathbf U\mathbf X^T((\widehat{\boldsymbol\mu}_1-\boldsymbol\mu_1)+(\widehat{\boldsymbol\mu}_2 -\boldsymbol\mu_2))}{\sqrt{(\widehat{\boldsymbol\mu}_1-\widehat{\boldsymbol\mu}_2)^T\mathbf X\mathbf U^T\Lambda^{-2}(\widehat{\boldsymbol\Sigma}_0)\mathbf U\mathbf X^T(\widehat{\boldsymbol\mu}_1-\widehat{\boldsymbol\mu}_2)}}.
\end{aligned}
\end{equation}
For the numerator, we have
\begin{equation}
\begin{aligned}
  &{(\widehat{\boldsymbol\mu}_1-\widehat{\boldsymbol\mu}_2)^T\mathbf X\mathbf U^T\Lambda^{-1}(\widehat{\boldsymbol\Sigma}_0)\mathbf U\mathbf X^T((\widehat{\boldsymbol\mu}_1-\boldsymbol\mu_1)+(\widehat{\boldsymbol\mu}_2 -\boldsymbol\mu_2))} \\
  = &(\widehat{\boldsymbol\mu}_1-{\boldsymbol\mu}_1)^T\mathbf X\mathbf U^T\Lambda^{-1}(\widehat{\boldsymbol\Sigma}_0)\mathbf U\mathbf X^T(\widehat{\boldsymbol\mu}_1-{\boldsymbol\mu}_1)
  -(\widehat{\boldsymbol\mu}_2-{\boldsymbol\mu}_2)^T\mathbf X\mathbf U^T\Lambda^{-1}(\widehat{\boldsymbol\Sigma}_0)\mathbf U\mathbf X^T(\widehat{\boldsymbol\mu}_2-{\boldsymbol\mu}_2) \\
  &+({\boldsymbol\mu}_1-{\boldsymbol\mu}_2)^T\mathbf X\mathbf U^T\Lambda^{-1}(\widehat{\boldsymbol\Sigma}_0)\mathbf U\mathbf X^T((\widehat{\boldsymbol\mu}_1-\boldsymbol\mu_1)+(\widehat{\boldsymbol\mu}_2 -\boldsymbol\mu_2))\\
  =&T_1-T_2+T_3.
  \end{aligned}
\end{equation}
Due to the normalization, we know that $\xi_1=\mathbf U\mathbf X^T(\widehat{\boldsymbol\mu}_1-\boldsymbol\mu_1)$ follows the multivariate Gaussian distribution $\mathcal N(0,\frac1n\mathbf I)$, with $n=N/2$ being the training data number per class. Then, by Lemma \ref{lm:minus1moment} and $\|\xi_1\|^2\overset{a.s.}\longrightarrow2\gamma$, we have
\begin{equation}
  T_1 = \xi_1^T\Lambda^{-1}(\widehat{\boldsymbol\Sigma}_0)\xi_1 = {\|\xi_1\|^2}\frac{\xi_1^T}{\|\xi_1\|}\Lambda^{-1}(\widehat{\boldsymbol\Sigma}_0)\frac{\xi_1}{\|\xi_1\|}\overset{a.s.}\longrightarrow\frac{2\gamma}{1-\gamma}.
\end{equation}
Similarly, letting $\xi_2=\mathbf U\mathbf X^T(\widehat{\boldsymbol\mu}_2-\boldsymbol\mu_2)$, the same argument gives $T_2\overset{a.s.}\longrightarrow\frac{2\gamma}{1-\gamma}$. Denoting $\xi_3=\Lambda^{-1}(\widehat{\boldsymbol\Sigma}_0)\mathbf U\mathbf X^T({\boldsymbol\mu}_1-{\boldsymbol\mu}_2)$ and recalling Lemma \ref{lm:minus1moment}, we have
\begin{equation}
\begin{aligned}\label{eq:xi_3_bound}
  \|\xi_3\|^2&=({\boldsymbol\mu}_1-{\boldsymbol\mu}_2)^T\mathbf X\mathbf U^T\Lambda^{-2}(\widehat{\boldsymbol\Sigma}_0)\mathbf U\mathbf X^T({\boldsymbol\mu}_1-{\boldsymbol\mu}_2)
  \overset{a.s.}\longrightarrow\frac{\|\mathbf X^T({\boldsymbol\mu}_1-{\boldsymbol\mu}_2)\|^2}{(1-\gamma)^3}<\infty.
\end{aligned}
\end{equation}
Then, since $\xi$ follows $\mathcal N(0,\frac1n\mathbf I)$ and $\xi_3$ has bounded entries due to (\ref{eq:xi_3_bound}), we have
\begin{equation}
    \xi_3^T\xi_1\overset{a.s.}\longrightarrow0.
\end{equation}
Similarly, $\xi_3^T\xi_2\overset{a.s.}\longrightarrow0$. Thus,
\begin{equation}
  T_3=\xi_3^T(\xi_1+\xi_2)\overset{a.s.}\longrightarrow0.
\end{equation}
Therefore, we have the numerator $T_1-T_2+T_3\overset{a.s.}\longrightarrow0$.

For the dominator, letting $\zeta=\mathbf U\mathbf X^T(\widehat{\boldsymbol\mu}_1-\widehat{\boldsymbol\mu}_2)$, we have
\begin{equation}
\begin{aligned}
  &{\sqrt{(\widehat{\boldsymbol\mu}_1-\widehat{\boldsymbol\mu}_2)^T\mathbf X\mathbf U^T\Lambda^{-2}(\widehat{\boldsymbol\Sigma}_0)\mathbf U\mathbf X^T(\widehat{\boldsymbol\mu}_1-\widehat{\boldsymbol\mu}_2)}} = \|\zeta\|\sqrt{\frac{\zeta^T}{\|\zeta\|}\Lambda^{-2}(\widehat{\boldsymbol\Sigma}_0)\frac{\zeta}{\|\zeta\|}}  \overset{a.s.}\longrightarrow\frac{\lim\|\zeta\|}{(1-\gamma)^{3/2}}.
  \end{aligned}
\end{equation}
Note that $\lim\|\zeta\|>0$, because $\widehat{\boldsymbol\mu}_1\ne\widehat{\boldsymbol\mu}_2$ almost surely. Thus, the dominator must be positive. Therefore, we have $T$ in (\ref{eq:T}) has limit $0$.

\section{Conclusion}
FLDA is an important statistical model in pattern recognition. The result obtain in this paper enriches the existing theory of FLDA, by showing that the generalization ability of FLDA is mainly determined by the dimensionality to training sample size ratio $\gamma=D/N$, given $D$ and $N$ are reasonably large and $N>D$. Important conclusions from this result include: 1) to ensure FLDA performing well, training sample size only needs to scale linearly with respect to data dimensionality, although a quadratic number of parameters are to be estimated in the sample covariance; and 2) the generalization ability of FLDA (with respect to the Bayes optimum) is independent of the spectral structure of the population covariance, given its nonsingularity and above conditions.


\bibliographystyle{IEEEtran}
\bibliography{gb_lda}
\end{document}